\documentclass[sn-mathphys-num]{sn-jnl}

\usepackage{graphicx}%
\usepackage{multirow}%
\usepackage{amsmath,amssymb,amsfonts}%
\usepackage{amsthm}%
\usepackage{mathrsfs}%
\usepackage[title]{appendix}%
\usepackage{xcolor}%
\usepackage{textcomp}%
\usepackage{manyfoot}%
\usepackage{booktabs}%
\usepackage{algorithm}%
\usepackage{algorithmicx}%
\usepackage{algpseudocode}%
\usepackage{listings}%
\usepackage{todonotes}
\usepackage{caption}
\usepackage{subcaption}
\usepackage[final]{changes}

\begin{document}

\title[Article Title]{Mixed Data Clustering Survey and Challenges}


\author*[2]{\fnm{Guillaume} \sur{Guerard}}\email{guillaume.guerard@devinci.fr}

\author[2]{\fnm{Sonia} \sur{Djebali}}\email{sonia.djebali@devinci.fr}

\affil*[2]{\orgname{Léonard de Vinci Pôle Universitaire, Research Center}, \orgaddress{\street{12 Avenue Léonard de Vinci}, \city{Paris La Défense}, \postcode{92916}, \country{France}}}


\abstract{The advent of the big data paradigm has revolutionized the way industries handle and analyze information, ushering in an era characterized by unprecedented volumes, velocities, and varieties of data. In this context, mixed data clustering emerges as a critical challenge, necessitating innovative approaches to effectively harness the wealth of heterogeneous data types, including numerical and categorical variables. Traditional methods, designed for homogeneous datasets, often fall short in accommodating the complexities introduced by mixed data, highlighting the need for novel clustering techniques tailored to this context. Hierarchical and explainable algorithms play a pivotal role in addressing these challenges, offering structured frameworks that enable interpretable clustering results, which are essential for informed decision‑making. This paper presents a method based on pretopological spaces. Moreover, benchmarking against traditional numerical clustering methods and pretopological approaches provides valuable insights into the performance and efficacy of our novel clustering algorithm within the big data paradigm.}




\maketitle

\section{Introduction}

The big data paradigm represents a seismic shift in the way industries approach and harness data, characterized by unprecedented volumes, velocities, and varieties of information \cite{patgiri2016big}. This paradigm challenges traditional data management and analysis techniques by demanding innovative solutions capable of processing, analyzing, and deriving insights from vast and diverse datasets. In particular, the inclusion of mixed data types, such as numerical and categorical variables, poses significant challenges to conventional methodologies, necessitating the development of novel approaches to effectively leverage the wealth of information available \cite{tsai2015big}.

Traditionally, data handling methods were designed around homogeneous datasets, typically consisting of numerical values. However, the big data paradigm introduces a multitude of data types, including structured, unstructured, and semi-structured data, which demand a departure from traditional approaches. Moreover, the three primary characteristics of big data—volume, velocity, and variety—amplify the complexity of data analysis, requiring scalable and adaptable solutions capable of processing large volumes of data at high speeds while accommodating diverse data formats and structures.

These methods for handling mixed data often involve separate analyses of categorical and numerical variables, treating them as distinct entities rather than integrating their interdependencies. While this approach may provide insights into individual data types, it fails to capture the inherent relationships and interactions between different variables, limiting the holistic understanding of the dataset. As such, there is a pressing need to bridge the gap between traditional methodologies and the complexities introduced by mixed data in the context of machine learning, especially clustering methods.

Understanding the limitations of clustering methods and identifying the gaps in current approaches is essential for advancing mixed data analysis. By critically assessing existing methodologies and their applicability to diverse datasets, we can pinpoint areas for improvement and develop innovative solutions tailored to the complexities of mixed data. Moreover, establishing a comprehensive understanding of traditional methods enables researchers to build upon existing knowledge and leverage insights from diverse disciplines to address emerging challenges effectively.

Considering the big data paradigm, there is a growing need for hierarchical and explainable algorithms for mixed data clustering in all sectors \cite{ahmad2019survey,yu2006kernel,han2021exploring,mcparland2017clustering,caruso2021cluster}. Hierarchical clustering offers a structured approach that aligns well with the complexities of mixed data, allowing for the identification of nested patterns and relationships within the dataset. Moreover, hierarchical clustering facilitates interpretability by organizing data into a hierarchical, tree‑like structure, enabling stakeholders to understand the underlying logic behind clustering decisions. In an era where transparency and accountability are paramount, explainable algorithms play a crucial role in fostering trust and confidence in the clustering process, especially in sensitive domains such as healthcare and finance. Therefore, the development of hierarchical and explainable algorithms tailored to mixed data clustering is essential for enabling meaningful insights and informed decision‑making in the era of big data.
Furthermore, evaluating the quality and significance of resulting clusters and deriving insights from them can be challenging because traditional clustering techniques designed for numerical data are not directly applicable to mixed data, as they often rely on a Euclidean space. Consequently, we will present less straightforward analyses of mixed data clusters using dimensionality reduction.

Addressing the remaining challenges in mixed data analysis requires a multifaceted approach that integrates advanced computational techniques with domain‑specific expertise. By leveraging pretopology \cite{dalud2001pretopology}, an emerging field that combines topological principles with data analysis, we can develop robust algorithms capable of building a hierarchical clustering of mixed datasets without the need for dimensionality reduction. By constructing a logical space that accounts for the inherent relationships between different data types, pretopology allows for the seamless integration of numerical, categorical, and temporal variables within a unified framework. This approach offers a promising avenue for addressing the complexities of mixed data analysis in the era of big data, providing interpretable and actionable insights for decision‑making across various industries. This method offers various advantages such as:

\begin{itemize}
\item No dimensionality reduction required
\item Customizable logical space creation
\item Hyperparameters for clustering and division conditions, allowing for tailored hierarchy creation
\item XAI on the dendrogram by comparing cluster characteristics grouped at each branching
\end{itemize}

The contributions of this paper are as follows:

\begin{enumerate}
\item a state‑of‑the‑art review of dimensionality reduction methods;
\item a state‑of‑the‑art review of mixed data clustering methods;
\item a state‑of‑the‑art review of clustering evaluation measures;
\item a new pretopology‑based clustering algorithm;
\item benchmarking of these methods on datasets, including a synthetic data generator;
\item an in-depth discussion of the remaining challenges in mixed data clustering in the era of big data.
\end{enumerate}

The paper is organized as follows: Section \ref{sec:methods} presents an overview of the benchmark. Dimensionality reduction methods are described in Section \ref{sec:dimred}, clustering algorithms in Section \ref{sec:alg}, and evaluation measures in Section \ref{sec:measures}. The datasets are presented in Section \ref{sec:mat}, and results are shown in Section \ref{sec:res}. A discussion of bottlenecks and challenges in mixed data is provided in Section \ref{sec:disc}, and we conclude in Section \ref{sec:ccl}.

\section{Methods}\label{sec:methods}

Research in mixed‑data clustering has primarily focused on modifying existing clustering algorithms originally designed for numerical or categorical data to handle mixed datasets effectively. Based on the survey by Ahmad et Al. \cite{ahmad2019survey}, four main clustering paradigms can be distinguished: \emph{partitional clustering}, \emph{hierarchical clustering}, \emph{model‑based clustering} and \emph{neural‑network‑based clustering}.

\emph{Partitional clustering} partitions the dataset into disjoint clusters and evaluates the partition using a cost function. Each cluster is represented by a centroid chosen to minimize the distance between its member points and the centroid relative to other centroids. Data points are iteratively reassigned until the cost function is minimized. This cost is typically the sum of distances between each data point and its nearest centroid.

\emph{Hierarchical clustering} builds a nested hierarchy of clusters with two main variants: agglomerative and divisive. In the agglomerative variant, each data point is initially treated as a separate cluster, a similarity matrix between clusters is computed, and the two most similar clusters are then merged. This process continues until a single cluster remains, producing a dendrogram. The divisive variant reverses this procedure, beginning with a single cluster of all data points and recursively splitting clusters. A desired number of clusters can be obtained by cutting the dendrogram at the chosen level.

\emph{Model‑based clustering} assumes that each data object is generated by an underlying model, typically a statistical distribution. Fitting this model to the observed data yields both the cluster definitions and the assignment of data points.

\emph{Neural‑network‑based clustering} methods typically employ deep neural networks to transform the input data into cluster‑friendly representations (\cite{aljalbout2018clustering,min2018survey,karim2021deep}). Architectures such as multilayer perceptrons, convolutional neural networks or generative adversarial networks are commonly used. Representations, or latent features, are extracted from one or more layers and used as input to a clustering algorithm. Typically, the loss function used for training combines a \emph{network loss} and a \emph{clustering loss}. The network loss enforces reconstruction or preservation of information during training, while the clustering loss encourages the latent space to form well‑separated clusters.

Mixed datasets may exhibit various characteristics. Because some algorithms do not handle mixed data natively, Section~\ref{sec:dimred} introduces several dimensionality‑reduction methods. Each of the following algorithms was implemented with a consistent input–output interface. The selected methods are available as GitHub repositories or packages that can be integrated into the pipeline. The following algorithms are implemented in Python or hosted in frequently updated GitHub repositories:

\begin{itemize}
    \item \textbf{Dimensionality reduction}
        \begin{itemize}
            \item Factorial Analysis of Mixed Data (FAMD), introduced by \cite{escofier1979traitement}, see section \ref{subsec: FAMD};
            \item Laplacian Eigenmaps, introduced by \cite{belkin2003laplacian}, see section \ref{subsec: laplacian};
            \item Uniform Manifold Approximation and Projection (UMAP), introduced by \cite{mcinnes2018umap}, see section \ref{subsec: UMAP};
            \item Pairwise Controlled Manifold Approximation and Projection (PaCMAP), introduced by \cite{wang2021understanding}, see section \ref{subsec: PaCMAP}.
        \end{itemize}
    \item \textbf{Partitional}
    \begin{itemize}
        \item K-prototypes, introduced by \citet{huang1997clustering}, see section \ref{subsec:Partitional clustering -- K-prototypes};
        \item Convex K-means also known as Modha--Splanger, introduced by \citet{modha2003feature}, see section \ref{subsec:Partitional clustering -- Convex K-Means};
    \end{itemize}
    \item \textbf{Model-based}
    \begin{itemize}
        \item KAy-means for MIxed LArge data (KAMILA), introduced by \citet{foss2016semiparametric}, see section \ref{subsec:Model-based clustering -- KAMILA};
        \item Model Based Clustering for Mixed Data (ClustMD), introduced by \citet{mcparland2016model}, see section \ref{subsec:Model-based clustering -- ClustMD};
        \item Mixed Dataset and Dataset with Missing Values (MixtComp), introduced by \citet{biernacki2016bigstat}, see section \ref{subsec:Model-based clustering -- MixtComp}.
    \end{itemize}
    \item \textbf{Hierarchical}
    \begin{itemize}
        \item Phillip and Ottaway, introduced by \citet{philip1983mixed}, see section \ref{subsec:Hierarchical clustering -- Phillip and Ottaway};
        \item HDBSCAN with dimensionality reduction (DenseClus), introduced by \citet{mcinnes2017accelerated}, see section \ref{subsec:Hierarchical Density-Based clustering -- DenseClus};
        \item Pretopology, introduced by \citet{levy2022hierarchical}, see section \ref{subsec:Hierarchical based clustering -- Pretopologic}.
    \end{itemize}
\end{itemize}

Source code for the implemented algorithms is available at the following GitHub repository\footnote{\url{https://github.com/ClementCornet/Clustering-Mixed-Data/}}.

\section{Dimensionality reduction}\label{sec:dimred}

To handle high‑dimensional mixed data and project it into a lower‑dimensional numerical space, dimensionality reduction techniques are required. These techniques are used for data preprocessing, evaluation, or result visualization.

\subsection{Factorial Analysis of Mixed Data}\label{subsec: FAMD}

FAMD is a factor analysis method for mixed datasets. It applies separate factorial analyses to two groups of features—numerical and categorical—and then combines the results.

Consider a dataset with $K$ numerical variables $k=1,...,K$ and $Q$ categorical variables $q=1,...,Q$. . For a principal component $z$, let $r(z,k)$ denote the correlation coefficient between $z$ and $k$ and $\eta^2(z,q)$ the squared correlation ratio between $z$ and $q$. The main steps of FAMD are:

\begin{enumerate}
    \item Split the data into two groups: numerical features and categorical features.
    \item Perform a Multiple Correspondence Analysis (MCA) on the categorical features to maximize $\sum_{k} r(z,k)$.
    \item Perform a Principal Component Analysis (PCA) on the numerical features to maximize $\sum_{q} \eta^2(z,q)$.
    \item Perform a global PCA on the results of the two previous analyses. The global objective of FAMD is to maximize:
    \begin{equation}
        \sum_{k} r(z,k) + \sum_{q} \eta^2(z,q)
        \label{famd-obj}
    \end{equation}
\end{enumerate}

\begin{figure}
     \centering
     \begin{subfigure}[b]{0.45\textwidth}
        \centering
        \includegraphics[width=\textwidth]{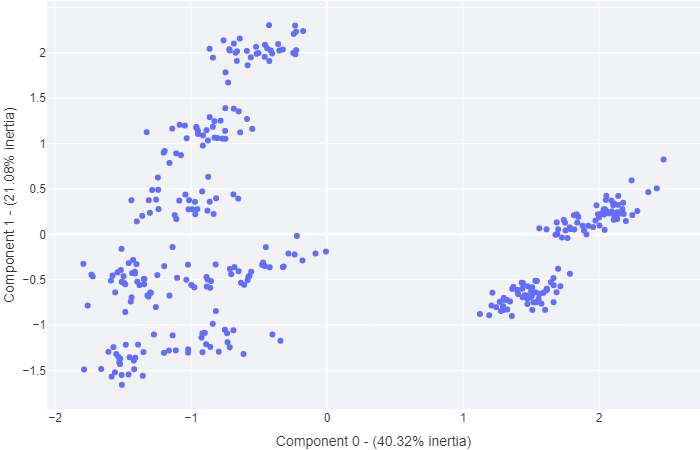}
        \caption{FAMD in 2D with explained inertia 61.40\%.}
        \label{famd-example}
     \end{subfigure}
     \hfill
     \begin{subfigure}[b]{0.45\textwidth}
        \centering
        \includegraphics[width=\textwidth]{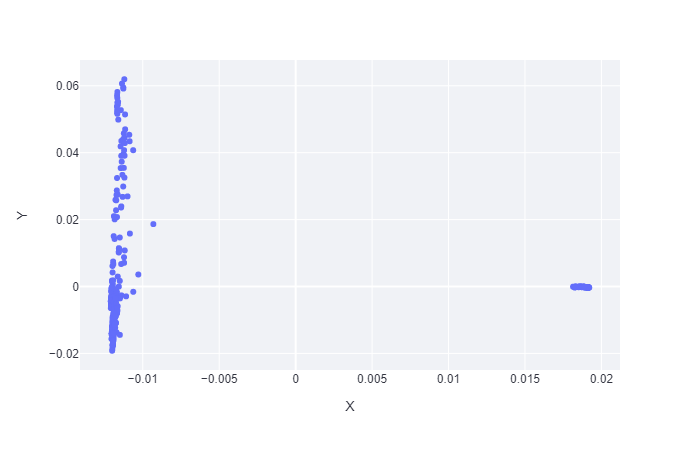}
        \caption{Laplacian Eigenmaps 2D with $t=1$.}
       \label{laplacian-example}
     \end{subfigure}
     \hfill
     \begin{subfigure}[b]{0.45\textwidth}
       \centering
       \includegraphics[width=\textwidth]{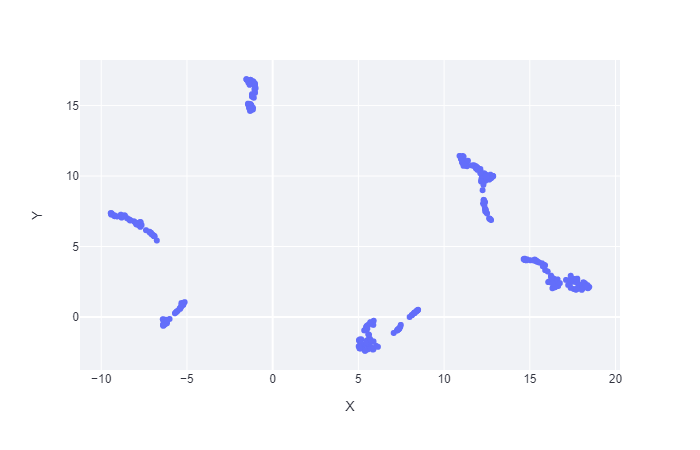}
       \caption{UMAP 2D with $k=15$.}
       \label{umap-example}
     \end{subfigure}
    \hfill
     \begin{subfigure}[b]{0.45\textwidth}
        \centering
       \includegraphics[width=\textwidth]{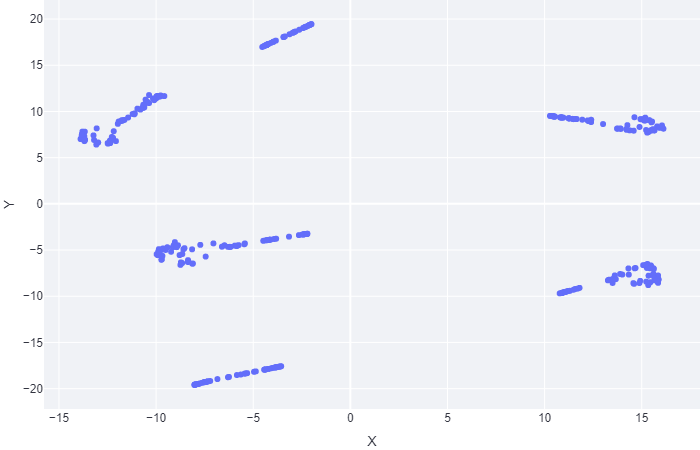}
       \caption{PaCMAP in 2 dimensions on the Palmer Penguins Dataset with FAMD initialization}
       \label{pacmap-example}
     \end{subfigure}
        \caption{Dimensionality reduction on Palmmer Penguins dataset.}
        \label{fig:three graphs}
\end{figure}

Explained inertia is the proportion of variance captured by the principal components, similar to explained variance in PCA. With FAMD, explained inertia is directly computed (as a factorial method) and no hyperparameter tuning is required, avoiding potential instability. However, FAMD may be limited when there are too few observations (leading to unstable MCA) or when the number of numerical features is much smaller than that of categorical features. An example of FAMD is shown in Figure \ref{famd-example}.

\subsection{Laplacian Eigenmaps} \label{subsec: laplacian}

Laplacian Eigenmaps is a spectral embedding technique used for non‑linear dimensionality reduction. The main steps of Laplacian Eigenmaps are as follows:
\begin{enumerate}
    \item Compute the pairwise distance matrix for the dataset. For mixed data, Huang’s distance is employed:
    \begin{equation}
    d_{ij} = d_{ij}^N + \gamma d_{ij}^C
    \label{huangdist}
    \end{equation}
    where $d_{ij}$ is the distance between two data points;  $d_{ij}^N$ is the squared Euclidean distance over numerical features; $d_{ij}^C$ is the Hamming distance over categorical features;  $\gamma$ is proportional to the average standard deviation of numerical features. Ratio is user defined, usually its half.
    \item Construct an adjacency matrix $W$ from the distance matrix. Although several methods exist, the Heat Kernel is most common: $W_{ij} = exp(-\frac{d_{ij}}{t})$ where $t$ is user-defined.
    \item Form the graph Laplacian. First, build the diagonal degree matrix $D$ representing the sum of the weights for every node, $D_{ii} = \sum_{j}W_{ji}$. The Laplacian matrix is $L = D - W$. Then, Solve the generalized eigenproblem $f$: $L f = \lambda D f$.
    \item Select the eigenvectors that define the low‑dimensional embedding. Since the first eigenvector corresponds to $\lambda=0$, the next $m$ eigenvectors are used to construct an $m$‑dimensional embedding.
\end{enumerate}

This approach is similar to spectral clustering and its results can be interpreted in a clustering context \cite{belkin2003laplacian}. However, adjacency computation requires hyperparameters (e.g., $t$), which can substantially affect the outcome. Moreover, unlike factorial methods, the axes in the spectral embedding lack explicit interpretation, leading to lower interpretability, as shown in Figure \ref{laplacian-example}.

\subsection{Uniform Manifold Approximation and Projection} \label{subsec: UMAP}

UMAP is a non‑linear dimensionality reduction algorithm that aims to preserve the local topological structure of the data. The algorithm first initializes an initial low‑dimensional embedding, then optimizes it via gradient descent. The process consists of the following steps:
\begin{enumerate}
    \item Compute the pairwise distance matrix for the dataset. For mixed data, we employ Huang’s distance (Equation \ref{huangdist}) \ref{huangdist}).
    \item Compute the adjacency matrix, representing the edges of a weighted graph. This requires a user‑defined hyperparameter $k$ (commonly set to 15). For each node, its $k$ nearest points are considered neighbors. The similarity $sim_{ij}$ between nodes $i$ and $j$ is computed using Equation \ref{umap-sim}.
    \begin{equation}
        sim_{ij} = exp(\frac{d_{ij} - d_N}{\sigma})
        \label{umap-sim}
    \end{equation}
    where $d_{ij}$ is the distance between $i$ and $j$; $d_N$ is the distance between $i$ and its nearest neighbor; $\sigma$ is adjusted for each node, so that the sum of weights for each node is $log_2(k)$. Because this similarity is not symmetric ($sim_{ij} \neq sim_{ji} $), the final weight stored in the adjacency matrix is $W_{ij} = (sim_{ij} + sim_{ji}) - sim_{ij}*sim_{ji}$.
    \item Initialize UMAP with a spectral embedding (e.g., Laplacian Eigenmaps) of this graph. Any dimensionality reduction technique can be used for initialization (even a random projection into a low‑dimensional space), but spectral embedding typically yields faster convergence.
    \item Optimization: For a data point $A$, randomly select one neighbor $N$ and one non‑neighbor $F$. For these two points, compute the low‑dimensional similarity score $s$ with $A$ as:
    \begin{equation}
        s = \frac{1}{1+\alpha d^{\beta}}
        \label{umap-lowd-sim}
    \end{equation}
    Here, $d$ is the distance in the low-dimension space, $\alpha=1.577 $ and $\beta=0.8951 $. \\
    From $s_{AN}$ and $s_{AF}$ the similarities between $A$ and $N$ and between $A$ and $F$, compute the cost function (\ref{umap-cost}):
    \begin{equation}
        cost = log(\frac{1}{s_{AN}}) - log(\frac{1}{ 1 - s_{AF}})
        \label{umap-cost}
    \end{equation}
    Then, $A$ is updated via stochastic gradient descent to minimize this cost and find its optimal position in the low‑dimensional embedding. 
\end{enumerate}

UMAP preserves the local structure of the data (illustrated in Figure \ref{umap-example}). This focus on local relationships leads to well‑defined clusters in the projected space. Indeed, UMAP can enhance the performance of numerical clustering algorithms \cite{allaoui2020considerably}. However, the number of neighbors $k$ significantly impacts the outcome and may bias interpretation.

\subsection{Pairwise Controlled Manifold Approximation and Projection}\label{subsec: PaCMAP} 

The PaCMAP method is very similar to UMAP. It also initializes a low-dimensional embedding and then optimizes it. However, PaCMAP aims to preserve both local and global structures, whereas UMAP focuses mainly on the local structure. PaCMAP’s main steps are:
\begin{enumerate}
    \item Initialization: With numerical-only data, Principal Component Analysis (PCA) is used to build the initial low-dimensional embedding. With mixed-data, we use FAMD, as it is considered the mixed-data counterpart of PCA. Note that, unlike UMAP, initialization affects the results of PaCMAP.
    \item Optimization: The following steps are repeated for $450$ iterations. PaCMAP relies on the concepts of Neighbors, Mid‑Near Pairs, and Further Pairs. For a given datapoint $A$, Neighbors are pairs formed by $A$ and its $k$ closest neighbors (hyperparameter, typically $10$). To define Mid‑Near Pairs, sample $6$ observations and select the pair of $A$ and the 2nd closest sampled observation. Further Pairs are pairs of $A$ and every other datapoint. The number of Mid‑Near and Further Pairs is given by hyperparameters, based on the number of Near‑Pairs.
    \item Construct a weighted graph for each iteration. The weights $w_{NB}$, $w_{MN}$ and $w_{FP}$ for Neighbors, Mid-Near Pairs and Further Pairs are assigned based on the pair type and the iteration:
    \begin{itemize}
        \item     - First 100 iterations : $w_{NB}=2$, $w_{MN}$ linearly decreases from 1000 to 3, $w_{FP}=1$; \\
    - Iteration 101 to 200 : $w_{NB}=3$, $w_{MN}=3$, $w_{FP}=1$; \\
    - Last 250 iterations : $w_{NB}=1$, $w_{MN}=0$, $w_{FP}=1$. \\
    \end{itemize}

    \item For each datapoint $i$, compute the loss function given by:
    \begin{equation}
    \begin{aligned}
        Loss = w_{NB} \times \sum_{J} \frac{\tilde{d_{ij}}}{10 + \tilde{d_{ij}}}  
        + w_{MN} \times \sum_{K} \frac{\tilde{d_{ij}}}{1000 + \tilde{d_{ik}}}  
        + w_{FP} \times \sum_{L} \frac{\tilde{d_{il}}}{1 + \tilde{d_{il}}}
        \label{pacmap-loss}
    \end{aligned}
    \end{equation}
    where $J$ is the set of neighbors $j$  of $i$; $K$ the mid-near points $k$  of $i$; $L$ the further points $l$ of $i$; $\tilde{d_{AB}} = 1 + ||y_A - y_B||^2$.
    \item Update i using stochastic gradient descent on the computed loss function (Equation \ref{pacmap-loss}) to find its optimal position in the low-dimensional space.
\end{enumerate}

The use of Mid‑Near Pairs enables PaCMAP to preserve the global structure of the data better than UMAP does. However, in some cases, preserving the global structure (rather than only the local structure) may provide no additional benefit. An example of PaCMAP is shown in Figure \ref{pacmap-example}.

\section{Algorithms}\label{sec:alg}

This section presents each implemented algorithm along with its hyperparameters, advantages, and drawbacks. The methods are summarized to provide a clear, consistent overview of their processes using uniform terminology and notation.

\subsection{Partitional clustering -- K-prototypes}
\label{subsec:Partitional clustering -- K-prototypes}

K‑prototypes is the most widely used partitional clustering algorithm for mixed datasets. It combines the numerical‑feature dissimilarity measure from K‑Means with the categorical‑feature dissimilarity measure from K‑Modes to define a mixed‑type dissimilarity (see Equation \ref{huangdist}). Each prototype merges a numerical centroid (mean) and a categorical mode.

Let $\gamma$ be a user-defined hyperparameter. It is a weight for categorical attributes in a cluster, in order to balance the influence of the two types of features. Let $k$ be a user-defined hyperparameter. 

The algorithm can be decomposed in three steps:

\begin{enumerate}
    \item Initialization of prototypes: For each of the $k$ clusters, select one data object as the initial prototype.
    \item Initial allocation: Assign each object to the cluster whose prototype yields the smallest dissimilarity. Update prototypes immediately after each assignment.
    \item Re‑allocation: For each object, recompute its dissimilarity to all prototypes. If a closer prototype is found, reassign the object and update the affected prototypes. Repeat this step until a full pass over the dataset produces no reassignments.
\end{enumerate}

A known limitation of this approach is that the Hamming‐based categorical distance treats all mismatches equally (0 or 1), potentially overlooking subtler relationships among categories. Prototype updates can also converge to local optima sensitive to initial selection.

Ahmad et Al. \cite{ahmad2007k} proposed enhancements to address these issues. Categorical dissimilarity is redefined using global frequency distributions and co‐occurrence statistics, rather than simple equality checks. Numerical features receive “significance” weights derived from discretizing each feature into intervals and evaluating their clustering impact. Although discretization informs the weight calculation, Euclidean distance continues to measure numerical dissimilarity. The resulting prototype retains the cluster mean for numerical attributes and represents categorical attributes by their proportional distributions within the cluster.

\subsection{Partitional clustering -- Convex K-Means}
\label{subsec:Partitional clustering -- Convex K-Means}

In Convex K-Means, given a dataset $S$ of $N$ data objects such that $S = (x_i; i = 1, ..., N)$, each data object $x_i$ is represented as a tuple of $M$ components of column vectors such that  $ x_i = (F_{(i,m)}; m = 1, ..., M)$ where $F_{m} = (F_{(i,m)}; i = 1, ..., N)$ is a column vector denoted as a \emph{feature vector}.   

Given a dataset, a feature space is defined by a set of features chosen from this dataset. This feature space contains all the values possible that the features of this set can take. The dimension of this space is equal to the number of features in the set.

Given a feature vector $F_{m}$, its components are all the values that the feature $m$ takes along the different data objects of $S$. For all $x_i$, the components $F_{(i,m)}$ lie in the same feature space \( \mathcal{F}_m \).

Given a data object $x_i = (F_{(i,1)}, F_{(i,2)}, ..., F_{(i,m)}, ..., F_{(i,M)})$, $x_i$ lies in a feature space \( \mathcal{F} \), created by the $M$-fold Cartesian product of the features spaces $\{\mathcal{F}_l\}^M_{m=1}$, such that: $ \mathcal{F} = \mathcal{F}_1 \times \mathcal{F}_2 \times ... \times \mathcal{F}_m \times ... \times \mathcal{F}_M  $. A feature vector $F_m$ can differ from the other feature vectors by its properties, especially by the type of the feature $m$. Then, each feature space $\mathcal{F}_m$ has its own properties (dimensions, topologies...) and is different from the others.

Given two data objects $ x_i = (F_{(i,m)}; m = 1, ..., M)$ and $x_j = (F_{(j,m)}; m = 1, ..., M)$, they propose a \emph{distortion measure} $D_m$ between the two corresponding feature vectors components $F_{(i,m)}$ and $F_{(j,m)}$. $D_m$ is assigned to the feature space $\mathcal{F}_m$ where $F_{(i,m)}$ and $F_{(j,m)}$ lie. From the $M$ distortion measures that can be obtained, they define a \emph{weighted distortion measure} $D^\alpha$ between $x_i$ and $x_j$ as a weighted sum of the $M$ distortion measures, such that : $D^\alpha(x_i, x_j) = \sum^{M}_{m = 1} \alpha_m D_m(F_{(i,m)}, F_{(j,m)})$. The features weighting is represented by the vector $\alpha = (\alpha_m; m = 1, ..., M)$, which contains the weights relative to each $D_m$. They are refered to as \emph{feature weights} and define the importance of a feature vector in the clustering. 

To adapt their algorithm to the mixed data case, they consider their dataset to have two feature spaces: one consisting of numerical features and the other consisting of categorical features. They represent a data object $x_i$ as a tuple of a  numerical feature vector component $F_{(i,1)}$ and a categorical feature vector component $F_{(i,2)}$, such that: $x_i = (F_{(i,1)}, F_{(i,2)})$. The distortion measures $D_1$ and $D_2$ are respectively the Euclidean distance and the cosine distance. 

Let the dataset be partitioned by the clusters $\{C_u\}^U_{u = 1}$. Given a cluster $C_u$, the cluster centroid is a tuple $c_u$ of $M$ components, such that : $c_u = (c_{(u,m)}; m = 1, ..., M)$. Along the features spaces $\{\mathcal{F}_m\}^M_{m=1}$, they denote the vector component $c_{(u,m)}$ as the centroid of the cluster $u$ lying in $\mathcal{F}_m$ with all the components of the feature vector $F_m$. The centroid $c_u$ is determined by the data object that minimizes the sum of the $D^\alpha$ between this data object and all the other data objects contained in $C_u$. To do so, each component $c_{(u,m)}$ is determined in the same way as $c_u$, but by minimizing the sum of the $D_m$ between the feature vector components, such that : $c_{(u,m)} = \underset{F_{(j,m)} \in \mathcal{F}_m}{\arg\min}(\sum_{x_i \in u} D_m(F_{(i,m)}, F_{(j,m)})) $.

They propose a method to automatically identify feature weights in order to reach a good discrimination between clusters along the features spaces $\{\mathcal{F}_m\}^M_{m=1}$. To do so, they define in the feature space $\mathcal{F}_m$ the \emph{average within-cluster distortion} denoted $\Gamma_m$ and the \emph{average between-cluster distortion} denoted $\Lambda_m$. A given features weighting $\alpha$ gives :
$\Gamma_m(\alpha) = \sum^U_{u=1} \sum_{x \in C_u}D_m(F_m, c_{(u,m)})$ where $ x = (F_{m}; m = 1, ..., M)$ and $\Lambda_m (\alpha) = \sum^N_{i = 1}D_m(F_{(i,m)}, \overline{c_m}) - \Gamma_m(\alpha)$ where $\overline{c} = (\overline{c_m}; m = 1, ..., M)$ denotes the generalized centroid of the dataset. The "best" $\alpha$ minimizes the $M$-product of the ratio between $\Gamma_m$ and $\Lambda_m$ and this optimal weighting scheme is found through an exhaustive grid-search. This is done by repeatedly running the algorithm with different features weightings over a fine grid on the interval $[0,1]$.

The algorithm can be decomposed into three steps:

\begin{enumerate}
    \item Start with an arbitrary partitioning by selecting initial centroids.
    \item Find the closest centroid for each data object using the proposed distortion measure.
    \item Compute the new centroids using the centroids definition mentioned above.
\end{enumerate}
\noindent Steps (2) and (3) are repeated until a stopping criterion is met.

The main drawback of this algorithm is its computational cost, which is high due to the brute-force search of feature weightings. The hyperparameters of the algorithm are the number of clusters to determine $k$ and the granularity of the exhaustive grid-search. Due to its limitations, Convex K-means does not meet our needs and often fails to provide satisfactory results on large datasets.

\subsection{Model-based clustering -- KAMILA} 
\label{subsec:Model-based clustering -- KAMILA}

KAMILA combines k-means clustering with the Gaussian–multinomial mixture model.

Parametric assumptions refer to how algorithms assume the data are “shaped.” For example, k-means clustering typically assumes that clusters’ shapes are spherical and that they are of similar size. Model-based clustering assumes that clusters’ shapes are defined by a given statistical distribution. Some parametric assumptions are more restrictive than others, and algorithm performance depends on both the strength of the assumptions and whether the data meet them.

Like k-means, KAMILA assumes that clusters of numerical data are spherical or elliptical, constituting relatively weak parametric assumptions. KAMILA also uses the properties of the Gaussian–multinomial mixture model \cite{hunt2011clustering} to balance the effects of numerical and categorical data without requiring the user to specify their relative weights.

Using KDE to estimate the mixture distribution of numerical data relaxes the Gaussian assumption. Kernel density estimation (KDE) is a nonparametric method for estimating a probability density function without information about the underlying distribution.

Let the dataset $S$ consists of $N$ observations, such that: $S = (X_i; i = 1, ..., N)$ where $X_i$ is the $i$-th observation. $P$ denotes the number of numerical features and $Q$ the number of categorical features. Each $X_i$ is a $(P+Q)$-dimensional vector of random variables $(V^T, W^T)^T$, such that : $X_i = (V_i^T, W_i^T)^T$ where $V = (V_i; i = 1,.., N)$ and $W = (W_i; i = 1,.., N)$. $V_i$ is a $P \times 1$ vector of numerical random variables and $W_i$ is a $Q \times 1$ vector of $ q = 1,2, ..., Q$ categorical random variables, such that : $W_i = (W_{i1}, ..., W_{iq}, ..., W_{iQ})^T$ where $W_{iq}$ is a categorical random variable that can have $L_q$ categorical levels, i.e the $L_q$ different categorical values that $W_{iq}$ can take, such that: $W_{iq} = \{1, ..., l, ..., L_q\}$. Then, a mixed data object $x_i$ is modeled as vector composed of a numerical part represented by a vector $v_i$ and a categorical part represented by a vector $w_i$, such that: $x_i = (v_i, w_i)$

Each $V_i$ follows a finite mixture of $G$ spherical or elliptical distributions (choice made by the user) such that in this case: $h_g(x;\alpha_g) = f_{V,g}(v_i; (\mu_g,\Sigma_g))$ where $\mu_g$ denotes the centroid of the $g$-th cluster and $\Sigma_g$ the scaling matrix of the $g$-th cluster.

Each $W_i$ follows a finite mixture of $G$ multinomial distributions such that in this case: $h_g(x;\alpha_g) = f_{W,g}(w_i; \theta_{g}) = \prod_{q = 1}^Q \eta(w_{iq}; \theta_{gq})$ where $\theta_{g} = (\theta_{qg}; q = 1, ..., Q)$, $\theta_{qg}$ denotes the parameters vector of the multinomial distribution corresponding to the $q$-th categorical variable contained in the cluster $g$ and $\eta$ is the multinomial mass function. $\theta_{qg}$ is a $L_q \times 1$ vector such that $\theta_{qg} = (\theta_{gql}; l = 1, ..., L_q)$. Each $\theta_{gql}$ is the probability that the $q$-th categorical variable  has the categorical level $l$ if the data object $x_i$ is in cluster $g$. The multinomial mass function is written as: 

\begin{equation}
    \eta(w_{q}; \theta_{gq}) = \prod_{l = 1}^{L_q} \theta_{gql}^{I\{w_{q} = l\}} 
    \label{MultMassFctn}
\end{equation}

\noindent where $I\{\cdot\}$ denotes the indicator function. 

Under the assumption that $V$ and $W$ are independent, the dataset $S$ follow a finite mixture of $G$ joint probability distributions of $(V^T, W^T)^T$ such that in this case: $h_g(x;\alpha_g) = f_{V,W,g}(v,w; (\mu_g, \Sigma_g, \theta_{g})) = f_{V,g}(v; (\mu_g,\Sigma_g)) \times f_{W,g}(w; \theta_{g})$.

We denote $\hat{\mu}_g$ the estimator of $\mu_g$ and $\hat{\theta}_{gq}$ the estimator of $\theta_{qg}$. The algorithm starts by initializing at iteration $t=0$ a set of centroids $\hat{\mu}^{(t)}_g$ and a set of parameters $\hat{\theta}^{(t)}_{gq}$. $\hat{\mu}^{(0)}_g$ can be initialized by random draws from an uniform distribution, but another work of \cite{foss2018kamila} specifies that random draws from the numerical variables of the observations give better results. $\hat{\theta}^{(0)}_{gq}$ is initialized by a random draw from a Dirichlet distribution. 

First comes the \emph{partition step}, which assigns each observation $i$ to a cluster $g$ according to the quantity $H_i^{(t)}(g)$. At the $t$-th iteration, with the set $\hat{\mu}^{(t)}_g$ and $\hat{\theta}^{(t)}_{gq}$, the assignment of an observation $i$ can be decomposed in 4 steps: 

\begin{enumerate}
    \item For the numerical features, the Euclidean distances $d_{ig}^{(t)}$ between $v_i$ and each $\hat{\mu}^{(t)}_g$ are computed before extracting the minimum distance. These two substeps are performed for the $N$ observations before obtaining the set $r^{(t)}$ of the $N$ minimum distances.
    
    \item $r^{(t)} $ is used to estimate $f_V$ through an univariate Kernel Density Estimation (KDE) step. KDE is a non-parametric estimation method used to estimate a density function of a random variable. This estimation is denoted $\hat{f_V}$.
    \item For the categorical features, the probability $f_{W,g}(w_i; \theta_{g})$ of observing $w_i$ in cluster $g$ is calculated.
    \item The function $H_i^{(t)}(g) = \log(f_{W,g}(w_i; \theta_{g})) + \log(\hat{f}_V(d_{ig}^{(t)}))$ is calculated. The observation $i$ is assigned to the cluster that maximizes $H_i^{(t)}(g)$.
\end{enumerate}

Then comes the \emph{estimation step}, where $\hat{\mu}^{(t+1)}_g$ and $\hat{\theta}^{(t+1)}_{gql}$ are calculated.  They are computed respectively as the mean of the numerical values in cluster $g$ over the number of data objects in $g$ and the mean of the number of occurrences of the categorical level $l$ in cluster $g$ over the number of data objects in $g$. The two estimators are then used as inputs for the partition step of the next iteration.

The process consisting of these two steps is repeated until a partition with stable clusters. Multiple runs of this process are performed with different initialization. At the final iteration of a given run, the sum over the N observations of the highest value of $H_i^{(final)}$ between the $G$ clusters. The algorithm outputs the partition generated by the run that maximizes the objective function. 

The hyperparameter of this algorithm is the number of runs to perform.

\subsection{Model-based clustering -- ClustMD}
\label{subsec:Model-based clustering -- ClustMD}

ClustMD uses a latent variable model (LVM). LVM's main idea is that the observed datapoints are correlated and forms particular patterns because they are influenced by hidden variables, called \emph{latent variables}.

 Let $S$ denote a dataset of $N$ observed data objects, such that $S = (x_i; i = 1,...,N)$. Each observed data object is a vector that contains $m$ mixed types variables (numerical, ordinal or categorical), such that: $x_i = ( x_{im}; m = 1, ..., M)$. The proposed model assumes that a given observed data object $x_i$ is the manifestation of an underlying latent numerical vector $z_i$, such that ${z}_i = (z_{im}; m = 1, ..., M)$. This representation enables to represent the different types of data with one unified type of variable. 
 
 The model proposes 3 ways to represent an observed datapoint regarding its type:

\begin{itemize}
    \item Case of numerical data: A given numerical variable $x_{im}$ is a numerical manifestation of a latent numerical variable $z_{im}$ that follows a Gaussian distribution. Both are of the same type, then : $x_{im} = z_{im} \sim \mathcal{N}(\mu_m, \sigma_m^2)$.
    
    \item Case of ordinal data: A given ordinal variable $x_{im}$ with $L_m$ levels is a categorical manifestation of a latent numerical variable $z_{im}$ following a Gaussian distribution, i.e $z_{im} \sim \mathcal{N}(\mu_m, \sigma_m^2)$. Both are of different type, so an adaptation is needed. Let $\gamma_m$ denotes a $L_m + 1$ vector of thresholds that partition the real line, such that: $\gamma_m = (\gamma_{(m,l)}; l = 1, ..., L_m)$. The observed datapoint $x_{im}$ is defined such that if $\gamma_{(j,l-1)} < z_{im} < \gamma_{(j,l)}$, then $x_{im} = l$. After this adaptation, $x_{im}$ is numerical so : $x_{im} = z_{im} \sim \mathcal{N}(\mu_m, \sigma_m^2)$.

    \item Case of categorical data: A given categorical variable $x_{im}$ with $L_m$ levels is a categorical manifestation of the components of a numerical latent vector $z_{im}$ of dimension $L_m - 1$. The vector $z_{im}$ follows a Multivariate Gaussian (MVN) distribution, i.e. $z_{im} = (z_{im}^l; l = 1, ..., L_m - 1) \sim \textbf{MVN}_{L_m - 1}(\underline{\mu}_m, \Sigma_m)$, where $\underline{\mu}_m$ is the mean vector and $\Sigma_m$ is the covariance matrix. The observed data object $x_{im}$ is defined such that :

    $$ x_{im} = \begin{cases} 1 &\text{if $\max_l\{z_{im}^l\} < 0$ ;}  \\ l &\text{if $ z_{im}^{l-1}=\max_{l}\{z_{im}^l\} $ and $z_{im}^{l-1} > 0$ for $l = 2, ..., L_m$}  \end{cases} $$
    
\end{itemize}

They represent the dataset as a matrix of $N$ rows and $M$ columns. Supposing that the numerical variables are in the first $C$ columns, the ordinal and binary variables in the following $O$ columns and the categorical data in the final $ M - (C + O)$ columns. Let $P = C + O + \sum^M_{m = C + O + 1} (L_m - 1)$, which is equal to the number of mixed type variables $M$. In clustMD, $z_i$  follows a mixture of $G$ multivariate Gaussian distributions of $P$ dimensions, i.e $z_i \sim \sum^G_{g=1} \pi_g \text{MVN}_P(\underline{\mu}_g, \Sigma_g$), where $\pi_g$ is the marginal probability of belonging to cluster $g$, $\underline{\mu}_g$ the mean for cluster $g$ and $\Sigma_g$ the covariance for cluster $g$.

The clustMD model is fitted, i.e obtaining the parameters of the statistical distributions in the mixture for which clustMD describes the best the observed data, using an Expectation-Maximization (EM) algorithm. EM is an iterative method used to find the maximum likelihood estimate of a latent variable, in our case $z_i$. The clustMD model derives firstly the complete data log-likelihood. Then, the Expectation step will compute the expectation of this complete data log-likelihood with respect to $z_i$. If categorical variables are present, a Monte Carlo approximation algorithm is used for the Expectation step. Finally, the Maximisation step will maximize the value of this expectation with regard to the model parameters.

\subsection{Model-based clustering -- MixtComp}
\label{subsec:Model-based clustering -- MixtComp}

This model-based clustering aims to cluster mixed dataset and dataset with missing values in a moderate dimensional setting. It is a statistical method for clustering mixed data, which combines the strengths of model-based clustering and Bayesian approaches. The method models mixed data as a mixture of multivariate distributions, with each component representing a cluster. It can handle different types of data, including continuous, discrete, and mixed data, as well as missing data. The method incorporates a latent variable model that captures the hidden structure of the data, enabling it to handle complex data structures. The clustering is performed through a Bayesian inference process, which estimates the number of clusters, cluster parameters and the latent variables that capture the underlying structure of the data.

Let the dataset $S$ consists of $N$ observations, such that: $S = (X_i; i = 1, ..., N)$ where $X_i$ is the $i$-th observation. One particular outcome of $X_i$ is the data object $x_i$, which has has $M$ different features, such that: $x_i = (x_{im}; m = 1, ..., M)$. A data object is decomposed in three parts: numerical, categorical and integer, such that: $x_i = (x_i^{\textbf{num}}, x_i^{\textbf{cat}}, x_i^{\textbf{int}})$. Each $x_{im}$ is contained in one of the three parts.
    
Each $X_i$ follows a finite mixture distribution of $G$ probability distributions such that: $h(x_i; \alpha_g) = f(x_{i}^{\textbf{num}}; \alpha_{g}^{\textbf{num}}) \times f(x_{i}^{\textbf{cat}}; \alpha_{g}^{\textbf{cat}}) \times f(x_{i}^{\textbf{int}}; \alpha_{g}^{\textbf{int}}) $ where $\alpha_g = (\alpha_{gm}; m = 1, ..., M).$ The density function $f$ is an univariate distribution associated to the feature $m$ if the data object $x_i$ is in cluster $g$.


The probability distribution of $g$ is chosen depending on the type of its corresponding feature $m$ :

\begin{itemize}
    \item Numerical type: the Gaussian model of \cite{celeux1995gaussian} is used.
    \item Categorical type: the multinomial model is used in the same way as KAMILA algorithm described in section \ref{subsec:Model-based clustering -- KAMILA}. In this case, let the data object $x_{im}$ have $L_m$ categorical levels, i.e  $x_{im} \in \{1, ..., l, ..., L_m\}$. Then, $f(x_{im}; \alpha_{gm}) = \eta(x_{im};\alpha_{gm})$ where $\alpha_{gm} = (\alpha_{gml}; l=1, ..., L_m)$ (see Equation \ref{MultMassFctn}).
    
    \item Integer type: the Poisson distribution of parameter $\alpha_{gm}$ is used, such that : $f(x_{im}; \alpha_{gm}) = \frac{(\alpha_{gm})^{x_{im}}e^{-\alpha_{gm}}}{{x_{im}!}}$.
\end{itemize}

\noindent To fit the model, MixtComp uses a variation of EM algorithm.


 \subsection{Hierarchical clustering -- Phillip and Ottaway}
 \label{subsec:Hierarchical clustering -- Phillip and Ottaway}

\cite{philip1983mixed} propose to use Gower's similarity measure to obtain a similarity matrix, which is then used as input for a hierarchical clustering algorithm. Gower's similarity measure separates categorical and numerical features into two subsets, creating one categorical feature space and one numerical feature space. In the categorical feature space, the similarity between two datapoints is computed by a weighted average of similarities between all categorical features, which is calculated using Hamming distance. In the numerical feature space, the similarity between two datapoints is computed by the sum of the similarities between all numeric features.

The equation for Gower's similarity measure is (by \cite{gower1971general}):

\begin{equation}
s_{ij} = \frac{\sum_{k=1}^{p}w_{ij}^{(k)}s_{ij}^{(k)}}{\sum_{k=1}^{p}w_{ij}^{(k)}}
\end{equation}

\noindent where $s_{ij}$ is the similarity between data points $i$ and $j$, $s_{ij}^{(k)}$ is the similarity between data points $i$ and $j$ for feature $k$, $p$ is the number of features. $w_{ij}^{(k)}$ is equal to 0 when $s_{ij}^{(k)}$ cannot be calculated because of missing values (or for other reasons).

If feature $k$ is categorical, then $s_{ij}^{(k)}$ is defined as:

\begin{equation}
s_{ij}^{(k)} = 
\begin{cases}
1 & \text{if data points $i$ and $j$ have the same value for feature $k$}, \\
0 & \text{otherwise}.
\end{cases}
\end{equation}
\\
If feature $k$ is numerical, then $s_{ij}^{(k)}$ is calculated as follows:
\begin{equation}
s_{ij}^{(k)} = \frac{|x_i^{(k)} - x_j^{(k)}|}{R_k}
\end{equation}

\noindent where $x_i^{(k)}$ and $x_j^{(k)}$ are the values of data points $i$ and $j$ for feature $k$, and $R_k$ is the range of values for feature $k$. 

\subsection{Hierarchical Density-Based clustering -- DenseClus}
\label{subsec:Hierarchical Density-Based clustering -- DenseClus}

Amazon proposes a python module named DenseClus\footnote{\url{https://github.com/awslabs/amazon-denseclus}}. This module performs a dimensionality reduction with UMAP method before using accelerated HDBSCAN algorithm from \cite{mcinnes2017accelerated}, an extension of HDBSCAN algorithm from \cite{campello2013density}.

HDBSCAN (Hierarchical Density-Based Spatial Clustering of Applications with Noise) is a hierarchical density-based clustering algorithm. A density-based clustering algorithm identifies contiguous regions of high density of objects in a data space, separated from other such clusters by contiguous regions of low density. The objects in the separating regions of low density are typically considered as noise/outliers (see \cite{sammut2011encyclopedia}).

Given a dataset $S$ of $N$ objects, such that $S = (x_i; i = 1,...,N)$, they define a \emph{core distance} of a data object $x_i$ with regard to the hyperparameter $k$, denoted $d_{\textbf{core}}(x_i)$, as the distance from this data object to its $k$-nearest neighbor, i.e the $k$-th data object closer to it. Core distance is smaller for a data object in a dense region of data objects, while sparser regions give larger core distances to objects. Core distance enables to estimate the density of a region, by taking the inverse of the core distance.

They also define a data object $x_i$ as an \emph{$\epsilon$-core object} for every value of the parameter $\epsilon$ that satisfies $d_{\textbf{core}}(x_i) \leq \epsilon$. This is equivalent to saying that the data object $x_i$ has its $k$-nearest neighbors in the neighborhood defined by $\epsilon$.
 
From the concept of the core distance, they define a new distance metric between two objects called \emph{mutual reachability distance}. Given two objects $x_i$ and $x_j$, the mutual reachability distance is computed as: 

$$d_{\textbf{mreach}}(x_i, x_j) = \textbf{max}\{d_{\textbf{core}}(x_i), d_{\textbf{core}}(x_j), d(x_i, x_j)\}$$ 

where $d(\cdot, \cdot)$ denotes a metric distance. The mutual reachability distance captures not only the distance between the two objects in the Euclidean space but also the density of their neighborhood.

They represent their data as a weighted graph called the \emph{Mutual Reachability Graph}. In this graph, the objects are considered to be the vertices. An edge between any two objects is considered to have a weight equal to the mutual reachability distance between the two objects. To model the cluster, all edges having weights greater than $\epsilon$ are removed and the remaining groups of connected $\epsilon$-core objects constitutes the clusters. The remaining unconnected objects are considered as "noise".

Clusters hierarchy is built with a divisive fashion (considering firstly all objects being contained in a single cluster) and by varying the value of $\epsilon$. After computing the core distance with regard to $k$ for all data objects in $S$, the algorithm computes the graph and extract the Minimum Spanning Tree (MST) from it using the Prim's algorithm. A MST is a subset of a graph that connects all the vertices of this graph together such that the sum of the edges weight is minimum. Then, it iteratively removes all edges from the MST in decreasing order of weights. This is done by sorting the edges of the MST in an increasing order and gradually decreasing the value $\epsilon$ so that a given edge with a weight above $\epsilon$ is removed. $\epsilon$ acts as a distance threshold, so that its variation gradually disconnect objects from their clusters. This is equivalent to gradually increasing a density threshold $\lambda = \frac{1}{\epsilon}$, so that a cluster not dense enough will be split. $\lambda$ is increased until no split is performed anymore.

Splits are not performed in a classical way but occurs under particular constraints. A \emph{minimum cluster size} parameter $\omega$ defines the minimum number of objects accepted in a cluster. When a parent cluster is split into two child clusters, if any of the two child cluster contains fewer objects than $\omega$, the split is considered as "spurious". The child cluster in question will be considered as "falling out of the parent cluster" at the given $\lambda$ value, labelled as "noise" and removed from the cluster. Three cases can be encountered after a cluster split:

\begin{enumerate}
    \item The two child clusters' sizes are below $\omega$. The child clusters are removed from the parent cluster. No other splits are executed after. 
    \item If only one child cluster's size is higher than $\omega$, it is considered as the continuation of the parent cluster and takes its parent cluster's label. The same cluster size evaluation process is repeated on it while the other child cluster is removed.
    \item If more than one child cluster contains more than $\omega$ data objects, the split is considered as "true". Two child clusters are obtained and the same cluster size evaluation process is repeated on them.
\end{enumerate}

We can consider that the parent cluster is "shrinking" through the splits of case (2), until case (1) or case (3) is encountered. 

From the obtained dendrogram, the clusters extraction is applied according to the \emph{stability} of the clusters, i.e their capacity to keep shrinking until a "true" split occurs as $\lambda$ increases. Let $S$ be partitioned by the clusters $\{C_u\}^U_{u = 1}$. Given a cluster $C_u$ and a data object $x$, they define $\lambda_{\textbf{min,}C_u}(x)$ as the minimum $\lambda$ value for which the data object $x$ is contained in $C_u$. In other words, $\lambda_{\textbf{min,}C_u}(x)$  is the value of $\lambda$ at which this cluster became a cluster of its own (after a split or from the root of the dendrogram). They define $\lambda_{\textbf{max,}C_u}(x)$ as the value of $\lambda$ when the data object $x$ falls out of cluster $C_u$.
Then, the stability of a cluster $C_u$, denoted $\sigma(C_u)$, is determined as: $\sigma(C_u) = \sum_{x \in C_u} \lambda_{\textbf{max,}C_u}(x) - \lambda_{\textbf{min,}C_u}(x)$. The partition of $U$ clusters that maximizes the score $\sum_{u \in U} \sigma(C_u)$ is selected under the following constraint: the partition cannot contain overlapping clusters. This is equivalent to the following condition: if a cluster is selected, its child clusters cannot be selected.

The algorithm has a quadratic complexity, which limits its applicability for large amount of data. To overcome this problem, \cite{mcinnes2017accelerated} proposes an accelerated version of HDBSCAN. In this algorithm, Prim's algorithm is replaced by the Dual Tree Boruvka algorithm proposed by \cite{march2010fast}, which is designed to determine MST in a metric space. Accelerated HDBSCAN adapted this algorithm to the mutual reachability distance and presents a log-linear complexity.

\subsection{Pretopology-based algorithm}
\label{subsec:Hierarchical based clustering -- Pretopologic}

Pretopology allows for the extraction, organization, and structuring of data into homogeneous groups, as well as the integration of multicriteria analysis (using quantitative data, qualitative data, and other types of characteristics describing complex systems, such as time series). Pretopology-based clustering exploits the logical construction of pretopological spaces to define the construction of hierarchical structures according to the similarity between elements on specific characteristics. Pretopology-based clustering and its application for clustering complex energy systems have been presented in \cite{levy2022hierarchical}.

A pretopological space is based on the concept of pseudoclosure: let $(U, a(.))$ be a tuple, where $U$ is a set of elements and $a(.)$ is a pseudoclosure function on $U$, constitutes a pretopological space.

We define a pseudoclosure function $a : \wp(U) \to \wp(U)$ on a set $U$, is a function such that:
 $a(\emptyset) = \emptyset$;
 $\forall A \mid A \subseteq U : A \subseteq a(A)$, 
where $\wp(U)$ is the power set of U.

The mathematical formalization of a pretopological space used in the clustering algorithm presented in this paper is based on three elements: 
\begin{itemize}
    \item A set of weighted directed graphs $G = {G_1 (V_1, E_1 ), G_2 (V_2, E_2 ), ..., G_n (V_n, E_n )}$,
    \item A set of thresholds $\Theta = {\theta_1, \theta_2, ..., \theta_n}$
    \item A boolean function $DNF (.) : (\wp(U), U) \to {True, False}$, expressed as a positive disjunctive normal form in terms of $n$ boolean functions $V_1 (A, x), ..., V_n (A, x)$, each associated with a graph, and whose truth value depends on the set $A$ and the item $x$.
\end{itemize}

We determine if an item $x \in U$ belongs to the pseudoclosure of a set $A$ in the following way:

\begin{itemize}
    \item $\forall V_i (A, x)$, $V_i (A, x) = True \Longleftrightarrow \sum_{e_{xy} \in G_i, y \in A} w(e_{xy} ) \ge \theta_i$, where $e_{xy}$ is the edge going from $x$ to $y$, and $w(e)$ is the weight of the edge $e$.
    \item The item $x \in U$ will belong to the pseudoclosure of $A$ $\Longleftrightarrow$ the $DNF (.)$ evaluates to True
\end{itemize}

This formalization was introduced in \cite{laborde2019pretopology}.

Exploiting the built preotopological space, the construction of a hierarchical clustering is applied following the following algorithm:

\begin{enumerate}
\item Determine a family of elementary subsets called seeds.
\item Construct the closures of the seeds by iterative application of the pseudoclosure function.
\item Construct the adjacency matrix representing the relations between all the identified subsets (even the intermediate ones).
\item Establish the quasi-hierarchy by applying the associated algorithm on the adjacency matrix.
\end{enumerate}

This pretopological-based clustering approach is being implemented in a Python library and can be applied simultaneously to various data types, making it a versatile and powerful clustering method.

\added{From this point, the remaining text for this subsection is new}

\paragraph{Comparison to other hierarchical methods}

This approach departs from existing topological techniques in three principal respects:
\begin{itemize}
    \item No global metric assumption: similarity is encoded locally via graph thresholds, accommodating mixed‑type data and heterogeneous scales.
    \item Adaptive neighborhoods emerge naturally from the pseudoclosure operator, capturing both local cohesion and larger‑scale connectivity without manifold‑learning parameters.
    \item Computational efficiency: closure expansion and adjacency‑matrix construction scale linearly in the number of edges and seeds, avoiding combinatorial complexity.
\end{itemize}

In contrast to Philip\&Ottoway’s bottom‑up agglomeration, which requires building a full Gower similarity matrix and incurs O(n²) time and storage under a global metric, PretopoMD constructs a thresholded adjacency graph and uses pseudoclosure operators to achieve near‑linear scaling without any global distance assumption. Unlike DIVCLUS‑T \cite{chavent2007divclus}, one‑feature‑at‑a‑time divisive splits (which yield axis‑aligned trees and often over‑partition noisy dimensions), PretopoMD’s closures emerge from the joint topology of all features, automatically capturing multi‑variable interactions without manual cut‑point heuristics. And rather than borrowing decision‑tree machinery as in Unsupervised Binary Trees \cite{ghattas2017clustering}, where splits are chosen by user‑specified purity or distance gains and stopping rules, the PretopoMD framework derives adaptive, seed‑specific neighborhoods directly from data connectivity, handling mixed‑type attributes and heterogeneous scales with no external split criteria or depth constraints. Main differences between those four algorithms are shown in Table \ref{tab:hier_vs_pretopomd}.

\begin{table}[ht]
\centering
\small
\begin{tabular}{|p{2cm}|p{2.1cm}|p{2.1cm}|p{2.1cm}|p{2.1cm}|}
\hline
Aspect & Philip \& Ottaway & DIVCLUS‑T & Unsupervised Binary Trees & PretopoMD \\
\hline
Mixed‐data support
& via Gower metric
& handles mixed natively
& handles mixed natively
& metric‐free; directly on raw features \\
\hline
Core principle
& metric + linkage
& single‐feature splits
& recursive binary splits
& pseudoclosure operators on thresholded graph \\
\hline
Hierarchy output
& full dendrogram
& balanced binary tree
& unbalanced binary tree
& parameterized closure lattice \\
\hline
Interpretability
& moderate (linkage paths)
& high (one feature per split)
& high (decision‑rule splits)
& medium; pseudoclosure radii need explanation \\
\hline
Parameters
& choice of linkage; cut‑height
& split‑stop criteria
& split‑stop / difficulty criteria
& graph thresholds; pseudoclosure radius \\
\hline
Metric dependence
& yes (Gower)
& no explicit global metric
& no explicit global metric
& no metric at all \\
\hline
Cluster‐shape flexibility
& arbitrary
& axis‐aligned
& axis‐aligned
& arbitrary via pseudoclosure \\
\hline
Pros
& well‐understood; mature
& very interpretable splits
& interpretable, fast
& native mixed‐data support; direct hierarchy \\
\hline
Cons
& quadratic; needs metric
& may over‐split; binary only
& depends on splitting rule
& threshold choices non‐standard \\
\hline
\end{tabular}
\caption{Comparison of hierarchical clustering methods for mixed data versus PretopoMD.}
\label{tab:hier_vs_pretopomd}
\end{table}

\subsection{In short}


Table \ref{inshort-algos} shows the characteristics of the different algorithms such as its type or the use of tandem analysis. An algorithm's ability to produce outliers, or to handle missing values might differentiate it from others. Also, algorithms needing a hyperparameter $k$ for the number of clusters to find must use the Elbow Method to find $k$, which could extend the computation time artificially. 

\begin{table}[!ht]
\begin{tabular}{|l|ccccc|}
\hline
Algorithm                & Type         & Needs K & Tandem                                                & Missing Values & Outliers \\ \hline
K-Prototypes             & Partitional  & Yes     & -                                                     & No             & No       \\
Modha-Spangler           & Partitional  & Yes     & -                                                     & Yes            & No        \\
Phillip \& Ottaway       & Hierarchical & Yes     & -                                                     & No             & No        \\
Kamila                   & Model-Based  & Yes     & -                                                     & No             & No        \\
ClustMD                  & Model-Based  & Yes     & -                                                     & No             & Yes        \\
MixtComp                 & Model-Based  & Yes     & -                                                     & Yes            & Yes        \\
DenseClus                & Hierarchical & No      & UMAP                                                  & No             & Yes      \\
Pretopological Algorithm & Hierarchical & No      & \multirow{3}{*}{\begin{tabular}[c]{@{}c@{}}FAMD\\ UMAP \\ PaCMAP\end{tabular}} & No             & Yes      \\
 & & & & & \\
 & & & & & \\\hline
\end{tabular}
\caption{Characteristics of the different algorithms used in this study}
\label{inshort-algos}
\end{table}


\section{Measures}\label{sec:measures}


To establish a benchmark, we need metrics. Some of those metrics are used to assess the cluster tendency of a dataset, while others are used to evaluate the result of a cluster analysis, see \citet{palacio2019evaluation}.

An important proportion of the datasets we use to compare the different algorithms have no feature considered as "true clusters", or this feature might not be relevant. Therefore, we do not focus on external indices that compare a clustering with "true clusters". We mainly use internal indices, that evaluate the quality of a partition.

One of the characteristics of this study is the use of mixed data. As we do not use numerical-only data, we cannot use traditional clustering evaluation indices without preprocessing, as they often require a Euclidean space to compute. To use them, we use dimension reduction techniques to translate our data into a Euclidean space, then compute evaluation indices in this space. \\

\subsection{Cluster tendency -- Hopkins Statistic}
To evaluate the results of a dimension reduction, or simply to discuss the cluster tendency of a dataset, we use the Hopkins Statistic from \cite{hopkins1954new}. It behaves like a statistical hypothesis test, where the null hypothesis is that the datapoints are uniformly distributed. To compute it on a set $X$ of $n$ points in $d$ dimensions:

\begin{itemize}
    \item Generate $\tilde{X}$, a random sample of $m \ll n$ datapoints from $X$. \cite{lawson1990new} suggests sampling 5\% of $X$.
    \item Generate $Y$, a set of $m$ randomly and uniformly distributed datapoints.
    \item Define $u_i$ the minimum distance of $y_i \in Y$ to its nearest neighbor in $X$.
    \item Define $w_i$ the minimum distance of $\tilde{x_i} \in \tilde{X}$ to its nearest neighbor in $X$.
\end{itemize}

Then compute $H$ the Hopkins Statisitc defined by: 

\begin{equation}
    H = \frac{\sum_{i=1}^{m} u_i^d}{\sum_{i=1}^{m} u_i^d + \sum_{i=1}^{m} w_i^d}
    \label{hopkins-stat}
\end{equation}

$H$ is bounded between 0 and 1. A value close to 1 indicates that the data has a high clustering tendency, its data points are typically much closer to other data points than to randomly generated ones. A value close to 0 indicates uniformly spaced data, and values around 0.5 indicate random data.
The Hopkins Statistic usually is a useful measure. However datasets with only one very dense cluster might obtain a high score, although running a cluster analysis over them would be pointless.

\subsection{Cluster tendency -- Improved Visual Assessment of Cluster Tendency}

In partitional clustering, the question of cluster tendency, i.e the number of clusters necessary to obtain a good partitioning, can have a high influence of the final performance of an algorithm. Usually, it is manifested by an hyperparameter $k$ inputted by the user before running the algorithm (e.g $k$-means, $k$-prototypes, ...). To address this question, \cite{havens2011efficient} propose the Improved Visual Assessment of Cluster Tendency (iVAT) algorithm. 

Given a dataset, a dissimilarity matrix can be computed. It is a square and symmetric matrix where each element represent the dissimilarity between two data objects of the dataset. Each element is scaled to the range $[0,1]$, the value $0$ describes the highest dissimilarity between two objects and the value $1$ the lowest. From this matrix, a visual interpretation can be extracted which is an image of grayscale pixels, where each pixel represent the dissimilarity between two objects. Each pixel's colors depends on the value of the corresponding dissimilarity, such that the darker a pixel is, the lower the dissimilarity value is. The image is characterized by a black diagonal of pixels, because each data object is exactly similar with itself.

iVAT will reorder this matrix in order to have a visualisation of the cluster tendency. Reordering is done in a way to have one or more dark blocks along the diagonal of the image. A potential cluster is represented by a dark block, which is a submatrix with low dissimilarities values. Objects that are members of a dark block are relatively similar to each other. Cluster tendency is determined by the number of black blocks along the image diagonal.

iVAT can be a good alternative to Elbow method, which can have decreased performannce in case of outliers in the dataset. However, iVAT is a visual method and the extraction of the number of cluster must be done by the user. Different viewers can have different interpretation of cluster tendency, especially in the case of unclear boundaries between the different dark blocks. To address this problem, \cite{wang2010ivat}  propose a similar algorithm named aVAT that uses some image processing techniques to determine automatically the number of cluster. Unfortunately, the source code is anavailable and the algorithm is not well documented.

In our benchmark, we use iVAT to determine the relevance of the computed Hopkins statistics on a dataset. Indeed, only knowing the clustering friendliness of a dataset through Hopkins Statistic can lead to a bad evaluation of the dataset. For example, a dataset with a good Hopkins statistic can present a cluster tendency of only one cluster through iVAT. In this case, clustering would be useless despite of the different interpreation we could have with Hopkins statistic.


\subsection{Cluster analysis -- Calinski-Harabasz} 

A standard index to evaluate the definition of clusters is the Calinski-Harabasz index from \cite{calinski1974dendrite}, also known as the Variance-Ratio Criterion. From a set of data points, and the result of a cluster analysis, we compute $s$ as described in Equation \ref{calinski-equation}. For a dataset $E$, with $n_E$ individuals, divided into $k$ clusters, the Calinski-Harabasz index is is the ratio of the sum of between-clusters dispersion
and of within-cluster dispersion for all clusters (respectively $B_k$ and $W_k$, defined in Equation \ref{calinski-intermediate}), where dispersion is defined as the sum of distances squared.
\begin{equation}
    W_k = \sum_{q=1}^{k} \sum_{x \in C_q} (x - c_q)(x-c_q)^T \ , \ 
    B_k = \sum_{q=1}^{k} n_q (c_q - c_E)(c_q - c_E)^T
\label{calinski-intermediate}
\end{equation}

\noindent with $C_q$ the set of $n_q$ points in a cluster $q$ of center $c_q$, and $c_E$ the center of $E$. The index $s$ is calculated by:

\begin{equation}
    s = \frac{tr(B_k)}{tr(W_k)} \times \frac{n_E - k}{k - 1}
\label{calinski-equation}
\end{equation}

This index returns a positive real number, where a higher Calinski-Harabasz score relates to a model with better-defined clusters.

\subsection{Cluster analysis -- Silhouette} 

The Silhouette Coefficient, from \cite{rousseeuw1987silhouettes}, also evaluates the definition of clusters. It is only computed using pairwise distances. Therefore, it is not only possible to use it along with dimension reduction techniques, but also with Huang's Distance (Equation \ref{huangdist}). A score is computed for each data point as described in Equation \ref{silhouette}, using $a$ the mean distance of a point with the other points of its cluster, and $b$ the mean distance with the points of the nearest cluster.

\begin{equation}
    silhouette = \frac{b-a}{max(a,b)}
    \label{silhouette}
\end{equation}

The Silhouette Coefficient of a set of points is the mean of the Silhouette Coefficient for each sample. It is bound between -1 for incorrect clustering, and +1 for highly dense clustering. A score of zero indicates that clusters are overlapping. \\

\subsection{Cluster analysis -- Davies-Bouldin}

To evaluate clusters separation, we use the Davies-Bouldin index from\cite{davies1979cluster}. For each pair of clusters $i$ and $j$, a similarity $R_{ij}$ is computed (Equation \ref{Rij-DBindex}). Then, the Davies-Bouldin index $DB$ is the mean of the highest similarities for each cluster (Equation \ref{DBindex-eq}).

\begin{equation}
    R_{ij} = \frac{s_i + s_j}{d_{ij}}
    \label{Rij-DBindex}
\end{equation}
\noindent where  $R_{ij}$ is the similarity between clusters $i$ and $j$; $s_i$ and $s_j$ are the average distances of points of clusters $i$ and $j$ to their centroids;
 $d_{ij}$ is the distance between the centroids of clusters $i$ and $j$.

\begin{equation}
    DB = \frac{1}{k} \sum_{i=1}^{k} \max_{i \neq j} R_{ij}
    \label{DBindex-eq}
\end{equation}

A low Davies-Bouldin index indicates well-separated clusters, where zero is the lowest possible score.

\subsection{In short}

\begin{table}[!ht]
\begin{tabular}{|l|l|l|}
\hline
Index & Advantages  & Drawbacks \\ 
\hline
Calinski-Harabasz & \begin{tabular}[c]{@{}l@{}}- Higher for well-defined clusters\\ - Widely used in the litterature\\ - Fast to compute\end{tabular}            & \begin{tabular}[c]{@{}l@{}}- Higher for convex clusters\\ than  other concepts of clusters \\ (i.e. Density Based)\\ - Needs a Euclidean Space\end{tabular} \\ \hline
Silhouette        & \begin{tabular}[c]{@{}l@{}}- Higher for well-defined clusters\\ - Does not require a Euclidean Space\\ - Bound between -1 and +1\end{tabular} & \begin{tabular}[c]{@{}l@{}}- Higher for convex clusters\\ - Results are often less eloquent \\ than other indices\end{tabular}                                            \\ \hline
Davies-Bouldin    & \begin{tabular}[c]{@{}l@{}}- Low when clusters are well separated\\ - Simple computation\end{tabular}                                        & \begin{tabular}[c]{@{}l@{}}- Higher for convex clusters \\ - Needs a Euclidean Space\end{tabular}                       \\ \hline
\end{tabular}
\caption{Advantages and Drawbacks of the different interval validation indices}
\label{tab-indices}
\end{table}
Table \ref{tab-indices} summarizes the advantages and drawbacks of the different internal validation indices used in this study. Those characteristics are given in terms of computation complexity, interpretability, and mathematical limitations.

\section{Materials \& Cluster tendency}\label{sec:mat}


To compare the results of different clustering algorithms, we need to establish a benchmark. Therefore, we use multiple datasets used in the literature (Palmer Penguins\footnote{\url{https://www.kaggle.com/datasets/parulpandey/palmer-archipelago-antarctica-penguin-data}}, Heart Failure\footnote{\url{https://www.kaggle.com/datasets/fedesoriano/heart-failure-prediction}}, 
Sponge\footnote{\url{https://archive.ics.uci.edu/ml/datasets/sponge}}). Additionally, we have implemented a dataset generator to compare the algorithms across as many configurations as possible.

\subsection{Benchmark datasets}

\begin{figure}[!ht]
    \centering
    \includegraphics[width=\textwidth]{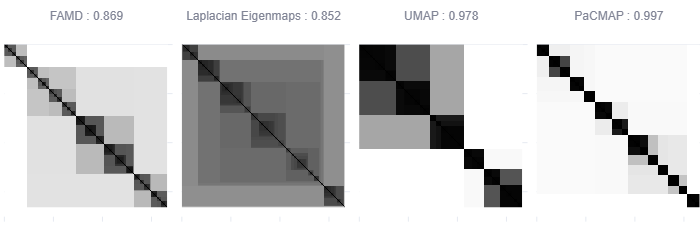}
    \caption{Hopkins Statistic and iVAT for every dimension reduction over the Palmer Penguins dataset.}
    \label{penguins-tendency}
\end{figure}

\paragraph{Palmer Penguins} The first results we present in this paper (others are available on the Github) are done on the Palmer Penguins dataset. This dataset is built upon physical measurements of 344 penguins in the Palmer Archipelago, in Antarctica \cite{gorman2014ecological}. It contains 4 numerical and 4 categorical features. We use it as a base case, as it is widely used in the literature, and its shape is pretty common. Also, it has high clustering tendency over the different dimension reductions (Figure \ref{penguins-tendency}), especially over UMAP and PaCMAP.

\paragraph{Heart Disease} The Heart Disease dataset belongs to the field of medicine. It combines 5 datasets over 13 features (5 numerical, 4 categorical, 4 ordinal). It contains 918 observations. Mixing in equal numbers each kind of features makes this dataset complex and the choose of metric or dimensionality reduction may completely change the clustering results.

\paragraph{Sponge} The Sponge dataset also belongs to the field of marine biology. Its aims is to describe and classify marine sponges. It has a pretty uncommon shape, as it only contains 75 individuals, with 42 categorical and 3 numerical features. Having both few individuals and a lot of categorical features makes this dataset harder to process, therefore interesting in the context of benchmarking.

Clustering tendency indicators for the Sponge and Heart Disease datasets are to be found on GitHub.

\subsection{Dataset generator}

To evaluate the different algorithms over every desired configurations, we use a dataset generator. The most common way to generate datasets to benchmark and evaluate clustering algorithms is to generate isotropic gaussian blobs. This method is natively present in the widely used scikit-learn for Python by \cite{scikit-learn}, MixSim for R  by \cite{melnykov2012mixsim} and Linfa for Rust\footnote{\url{https://rust-ml.github.io/linfa/}}. 

First, we generated cluster centers, with an average pairwise distance of 1. Then we generate samples from a gaussian mixture model with the density described by:

\begin{equation}
    p(x) = \frac{1}{k} \sum_{i=1}^{k} \mathcal{N}(\mu_i, \Sigma_i)
    \label{generator-density}
\end{equation}
where: \\
    - $k$ is the number of clusters \\
    - $\mu_i$ are the cluster centers \\
    - $\Sigma_i$ refers to the cluster covariances. Here, it is a diagonal matrix of the clusters variance.  
    
Inspired by \cite{costa2022benchmarking}, we split features upon quantiles to transform them into categorical features. Thus, we get a mixed dataset. With this method, the different parameters we can tune to obtain different configurations are:
\begin{itemize}
    \item The number of samples to generate (the number of individuals);
    \item The number of clusters $k$;
    \item The number of numerical features;
    \item The number of categorical features;
    \item The number of unique values taken by categorical variables;
    \item The standard deviation of clusters.
\end{itemize}

\section{Clustering Results and Discussions}\label{sec:res}

For the results, we will provide the analysis over the Penguins dataset and the dataset generator. More results are available on the Github\footnote{\url{https://github.com/ClementCornet/Benchmark-Mixed-Clustering}}.

\subsection{Computation cost and technical limitations}

Clustering algorithms are generally computation-heavy. Their respective computation times and memory usage should not be neglected, as they could cause technical limitations. The following execution statistics are obtained upon testing on a configuration with an AMD Ryzen 7 5800H CPU, on a 3.20GHz frequency with 512KB of L1 cache and 32GB of DDR4 RAM.

\begin{figure}[!ht]
    \centering
    \includegraphics[width=\textwidth]{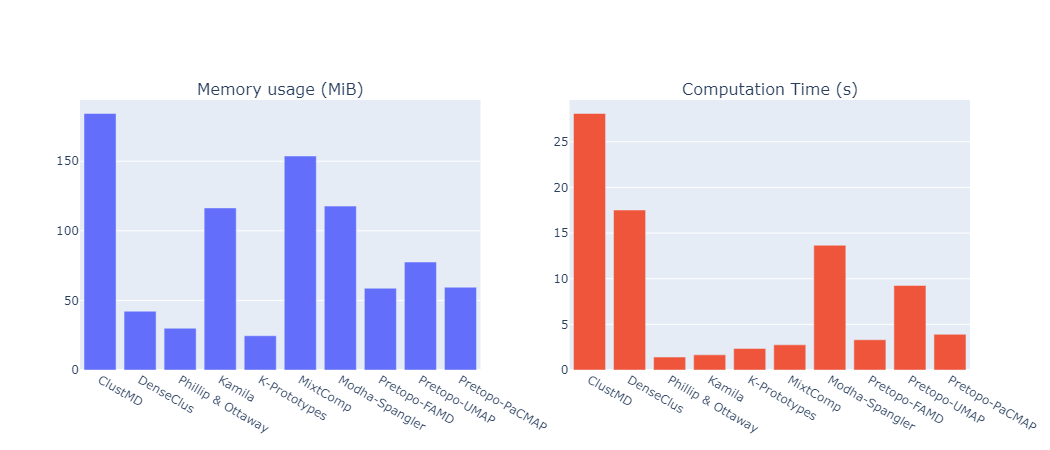}
    \caption{Time and Memory usage of the different algorithms, on a base case with 500 individuals, 5 numerical and 5 categorical features.}
    \label{base-config-cost}
\end{figure}

To benchmark the memory usage and computation time of the different algorithms, we measure those indices over several different generated datasets. The aim is to determine the impact of the dataset characteristics (number of individuals, number of numerical and categorical features) on its computation cost. To do so, we start from a "base configuration" (Figure \ref{base-config-cost}) with 500 individuals, 5 numerical and 5 categorical features. Then, we evaluate the impact of those 3 characteristics on the memory usage and computation time. Our measurements only include the clustering algorithm (not the data generation phase). We then measure the memory usage of this algorithm every 10 times/second, and keep the maximum.

\subsubsection{Number of Individuals}
\begin{figure}[!ht]
    \centering
    \includegraphics[width=\textwidth]{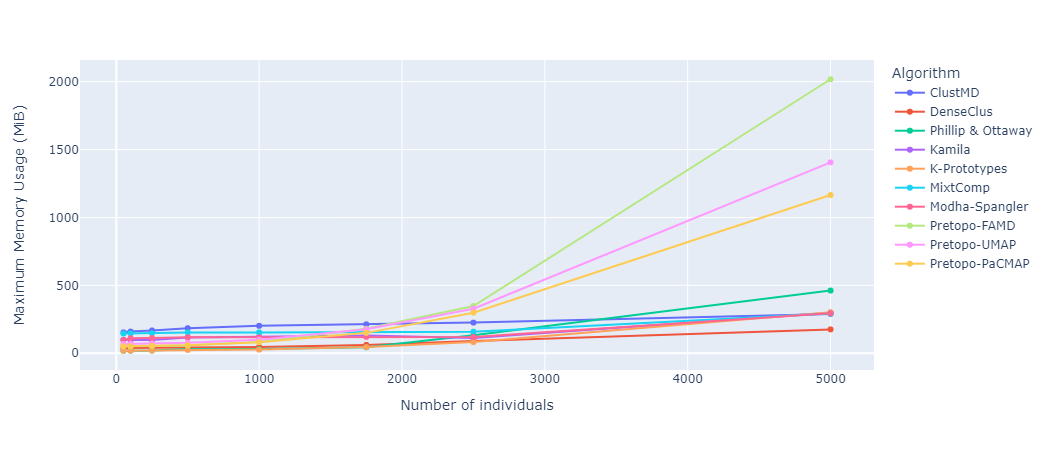}
    \caption{Maximum memory usage depending on the number of individuals}
    \label{indiv-mem}
\end{figure}

First, we evaluate the impact of the number of individuals on the computation time and memory usage. We include configurations with 50, 100, 250, 500, 1000, 1750, 2500 and 5000 individuals. From Figure \ref{indiv-mem}, we note that we have significative differences between the algorithms. The different variations of the pretopological algorithm have a steep curve, meaning that their memory usage increases faster than the other algorithms. On the other hand, we may note that most of the algorithms have similar memory usages with 5000 individuals.

\begin{figure}[!ht]
    \centering
    \includegraphics[width=\textwidth]{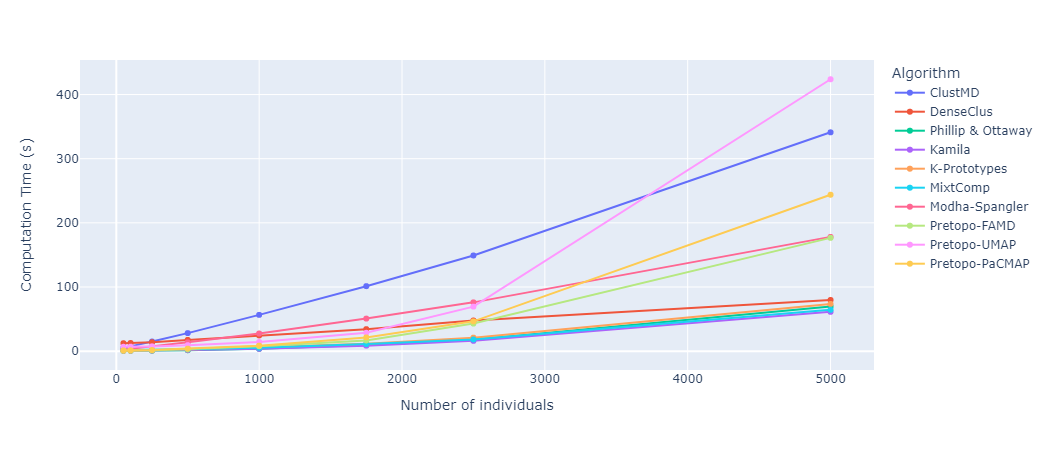}
    \caption{Computation time depending on the number of individuals}
    \label{indiv-time}
\end{figure}

Concerning the computation time (Figure \ref{indiv-time}), the pretopological algorithms' curves are closer to being linear, even if steeper than most algorithms. Yet the UMAP version takes 6 times more time with 5000 than with 2500 individuals. We may also note that ClustMD obtains very high computation time that may cause technical limitations, even if it seems linearly related to the number of individuals.

\subsubsection{Number of dimensions}

Then, we must evaluate how the number of dimensions impact the computation time and memory usage of the algorithms. As some algorithms treat numerical and categorical features in a totally different fashion, we evaluate their respective impacts separately. We measure the computation time and memory usage on generated datasets with 2, 5, 10, 20, 50 and 100 numerical/categorical features (depending on the characteristic we evaluate).

\begin{figure}[!ht]
    \centering
    \includegraphics[width=\textwidth]{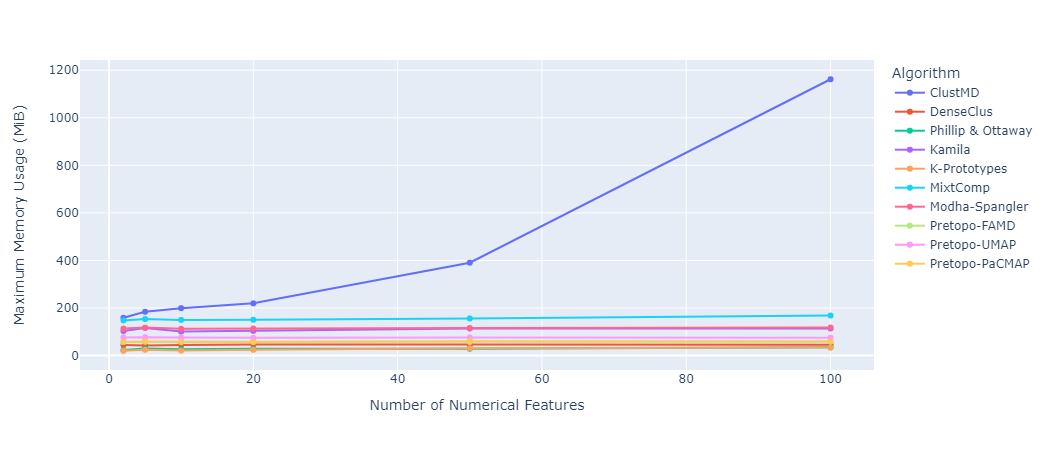}
    \caption{Maximum memory usage depending on the number numerical features.}
    \label{num-mem}
\end{figure}

\paragraph{Number of Numerical Features}
The number of numerical features seemingly has less impact on memory usage than the number of individuals (Figure \ref{num-mem}). Most algorithms barely use more memory with 100 numerical dimensions than with 2, so their results in terms of memory stay close to the base case. However, ClustMD's results are close to quadratic, and could cause limitations.

\begin{figure}[!ht]
    \centering
    \includegraphics[width=\textwidth]{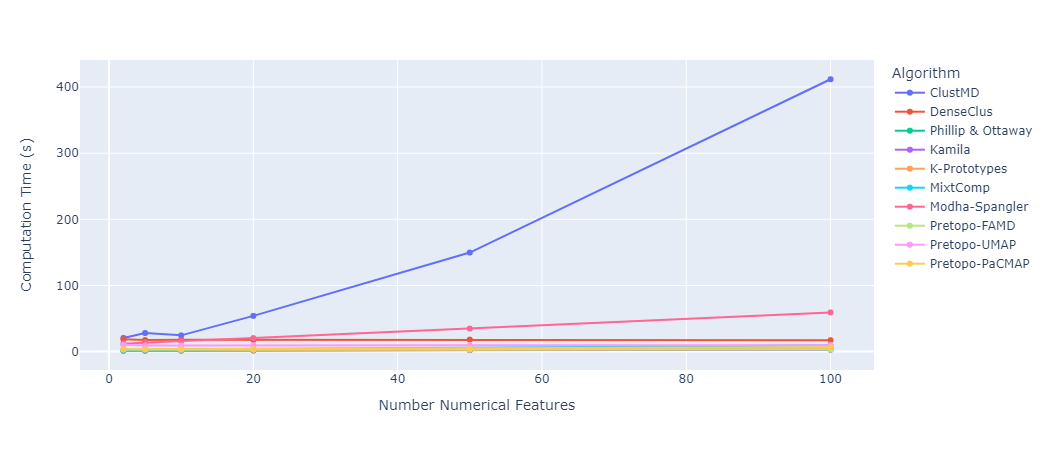}
    \caption{Computation time depending on the number numerical features.}
    \label{num-time}
\end{figure}

Observing the execution time leads to similar results than the memory usage (Figure \ref{num-time}). The number of numerical seems to have close to no impact there, except on ClustMD. We might also note that Modha-Spangler's execution time also increases slightly.

\paragraph{Number of Categorical Features}

\begin{figure}[!ht]
    \centering
    \includegraphics[width=\textwidth]{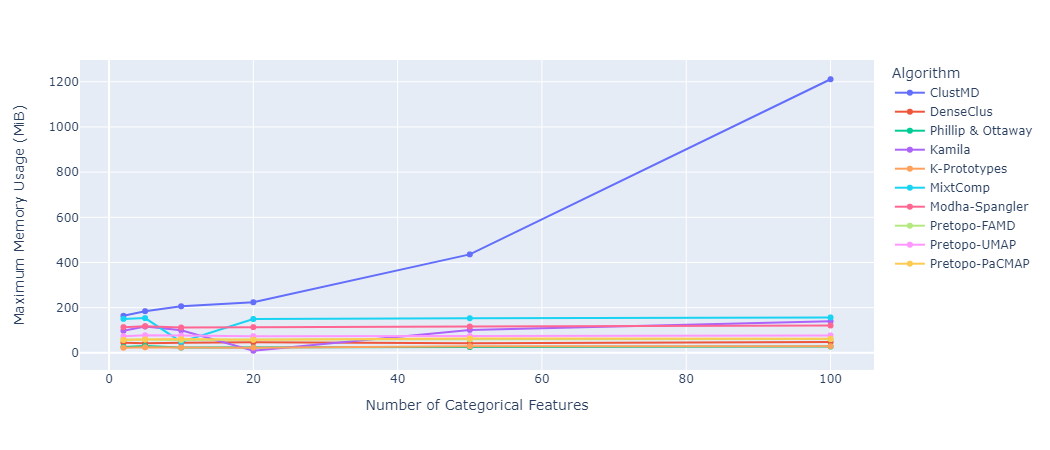}
    \caption{Maximum memory usage depending on the number categorical features.}
    \label{cat-mem}
\end{figure}

Measuring the memory usage of the algorithms over datasets over the number of categorical features leads to results very close to numerical features' ones, for every algorithm (Figure \ref{cat-mem}). 

\begin{figure}[!ht]
    \centering
    \includegraphics[width=\textwidth]{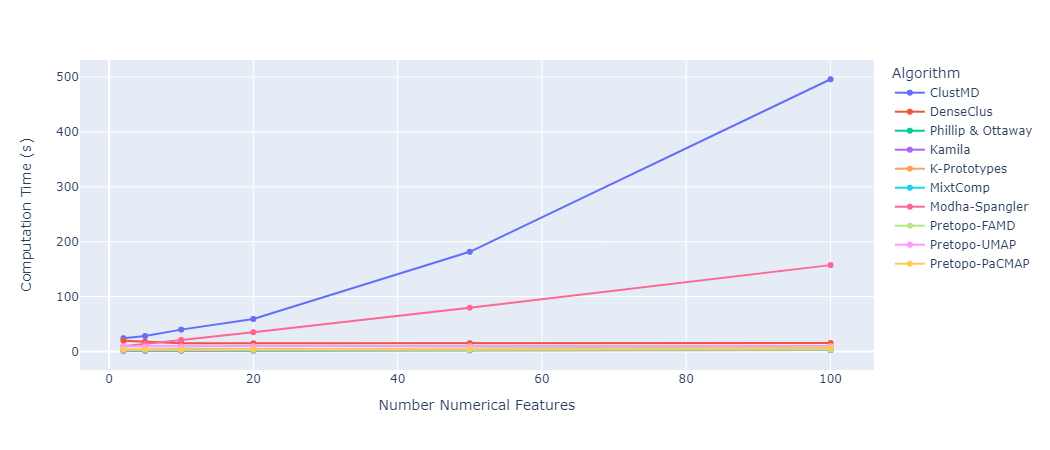}
    \caption{Computation time depending on the number categorical features}
    \label{cat-time}
\end{figure}

In terms of computation time (Figure \ref{cat-time}), the main difference in the impact of the number of categorical and numerical features can be observed in MixtComp. Its computation time is close to linearly related to the number of categorical features, while it was increasing slower upon the number of numerical features.

\subsubsection{Discussion}

\begin{figure}[!ht]
    \centering
    \includegraphics[width=\textwidth]{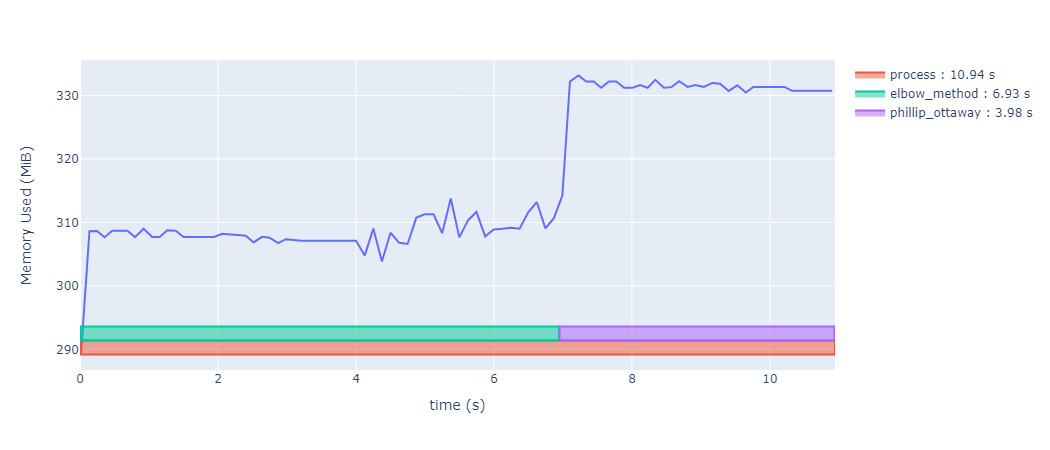}
    \caption{Impact of the elbow method on the computation cost.}
    \label{elbow-cost}
\end{figure}

\paragraph{Determining the number of clusters}
A large proportion of the algorithms need $K$ the number of clusters as a parameter. When $K$ is not known, we use the Elbow Method to determine it. As recommended, Elbow with mixed data is a combination of Gower distance with Calinski-Harabasz metric for each value of $K$ on a K-Means algorithm. However, the Elbow Method takes time, sometimes even more than the proper clustering. Figure \ref{elbow-cost} shows the memory usage over time in the different phases of Phillip \& Ottaway's algorithm, on the generated dataset with 1000 individual and 50 features of each type. The Elbow Method does not use more memory than the proper clustering, but more than twice the time. Yet, every algorithm needing $K$ as a parameter would take at least this time to process.

\begin{figure}[!ht]
    \centering
    \includegraphics[width=\textwidth]{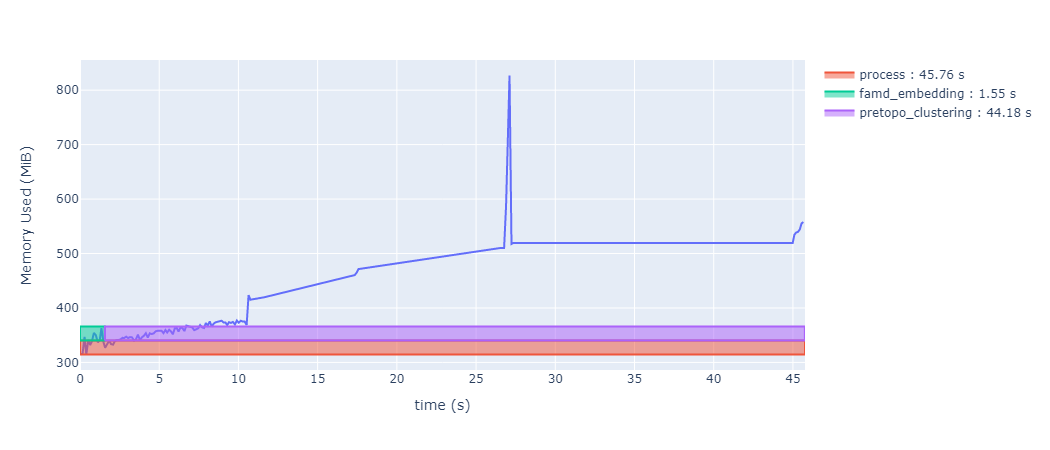}
    \caption{Details of the memory usage of the pretopological algorithms.}
    \label{pretopo-mem}
\end{figure}

\paragraph{Memory Usage of the Pretopological algorithm}
Our 3 versions of the pretopological algorithm rely on the same clustering techniques, but with a different tandem analysis. Also, their respective memory usage increases faster than other algorithms when the number of individuals increases (Figure \ref{indiv-mem}). However, on datasets with a large number of individuals, the pretopological algorithms encounter a brief peak of memory usage, when the memory usage sometimes doubles for 0.1 or 0.2s. Figure \ref{pretopo-mem} shows this phenomenon with the FAMD version, on a dataset with 2500 individuals, and 100 dimensions of each type.

\subsection{Dimensionality Reduction Results}
\added{New subsection}

Four dimensionality‐reduction methods—FAMD, Laplacian Eigenmaps, UMAP (with Huang’s mixed‐type distance) and PaCMAP—were applied to both the Palmer Penguins dataset and a synthetic high‐dimensional benchmark (500 points, 300 features, ten Gaussian clusters). Two complementary assessments were performed: cluster tendency in the low‐dimensional embeddings, and the quality of K‑Means clustering on those embeddings.

\begin{figure}[!ht]
\centering
\includegraphics[width=\textwidth]{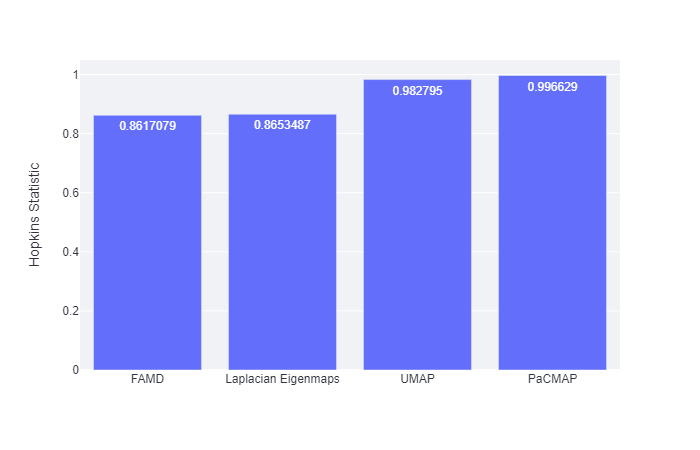}
\caption{Hopkins statistic for each dimensionality‐reduction technique on the Palmer Penguins dataset (average of 250 runs).}
\label{hopkins-penguins}
\end{figure}

\begin{figure}[!ht]
\centering
\includegraphics[width=\textwidth]{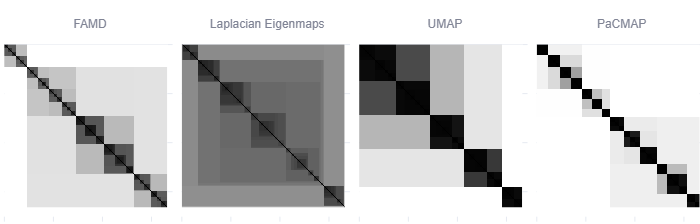}
\caption{iVAT visualizations of the four embeddings on the Palmer Penguins dataset. Clearer block‐diagonal patterns correspond to stronger cluster tendency.}
\label{ivat-penguins}
\end{figure}

First, cluster tendency was quantified by the Hopkins statistic (higher values indicate stronger departure from spatial randomness) and by iVAT (improved Visual Assessment of cluster Tendency), which reorders the pairwise distance matrix to reveal block‐diagonal patterns. Figure \ref{hopkins-penguins} presents the mean Hopkins statistic over 250 replications on the Palmer Penguins data, and Figure \ref{ivat-penguins} displays representative iVAT images. UMAP and PaCMAP attain the highest Hopkins scores and produce the most pronounced block structures in iVAT, suggesting that their embeddings preserve inherent cluster structure more effectively than FAMD or Laplacian Eigenmaps.

Second, K‑Means clustering was applied to each 2D embedding over a range of cluster counts, and the Davies–Bouldin index was recorded. Table \ref{dimr-penguins} reports the average DB on the Palmer Penguins data for $k=2,4,6,8$ (the best DB per method is highlighted). UMAP consistently yields the lowest DB, indicating more compact and well‐separated clusters. Laplacian Eigenmaps and FAMD attain intermediate DB values, while PaCMAP—despite strong tendency metrics—produces less compact clusters in this relatively small, noisy dataset.

\begin{table}[!ht]
\centering
\begin{tabular}{|c|cccc|}
\hline
Number of Clusters & FAMD & Laplacian Eigenmaps & UMAP & PaCMAP \\ \hline
2 & 1.10 & 1.12 & \textbf{0.74} & 1.68 \\
4 & 2.43 & 0.98 & \textbf{0.35} & 7.72 \\
6 & 3.50 & 1.18 & \textbf{0.31} & 11.14 \\
8 & 2.65 & 0.86 & \textbf{0.41} & 8.49 \\ \hline
\end{tabular}
\caption{Davies–Bouldin index for K‑Means on each 2D embedding of the Palmer Penguins dataset (averaged over 4 runs).}
\label{dimr-penguins}
\end{table}

On the synthetic benchmark of 500 points in 300 mixed‐type dimensions (ten Gaussian clusters), Table \ref{dimr-generated} reports the DB  for k=8–11. UMAP consistently achieves the lowest DB, demonstrating its ability to produce compact, well‐separated clusters even in very high‐dimensional embeddings. PaCMAP closely approaches UMAP’s performance and in fact attains a marginally better DB at k=10, reflecting its balance of local and global structure preservation in 300‑dimensional space. In contrast, FAMD and Laplacian Eigenmaps yield higher DB values that grow with increasing k, indicating diminished cluster separation as both dimensionality and cluster count rise.

\begin{table}[!ht]
\centering
\begin{tabular}{|c|cccc|}
\hline
Number of Clusters & FAMD & Laplacian Eigenmaps & UMAP & PaCMAP \\ \hline
8 & 5.81 & 1.69 & \textbf{0.75} & 0.79 \\
9 & 5.57 & 1.59 & \textbf{0.67} & 0.72 \\
10 & 5.18 & 1.75 & \textbf{0.52} & 0.65 \\
11 & 5.09 & 2.04 & \textbf{0.72} & 1.04 \\ \hline
\end{tabular}
\caption{Davies–Bouldin index for K‑Means on each 2D embedding of the synthetic high‐dimensional dataset (500 points in 300 mixed‐type dimensions).}
\label{dimr-generated}
\end{table}

In summary, UMAP with Huang’s mixed‐type distance provides the most compact, well‐separated embeddings for mixed‐type data in both settings, while PaCMAP may be preferred in very high‐dimensional regimes with clearly defined cluster boundaries.

\subsection{Clustering Results}

With the discussed materials and measures, we are able to evaluate the results of the different clustering algorithms.

In order to compute the Calinski-Harabasz, Davies-Bouldin and Silhouette scores, we translate datasets into Euclidean spaces using FAMD, with the output space having the same number of dimensions as the initial space. Here, FAMD is chosen over other techniques for the following reasons:
\begin{itemize}
    \item It is a factorial method, the inertia of the model is known
    \item It is deterministic
    \item It does not rely heavily on hyper-parameters
\end{itemize}
Also, as the Silhouette score is the only index of the study that can take a pairwise distance matrix as an input, we compute it with the Gower matrix. It might avoid a bias towards FAMD, or just add to the analysis in cases FAMD obtains low inertia.
Also, note that in some cases an algorithm may return only one cluster, or only outliers. In those cases, we display "-" in the results table. The following results come from the Penguins dataset, more results are available on the Github.

\paragraph{Palmer Penguins}

\added{Reworked}

All classical mixed‑data algorithms (Modha–Spangler, KAMILA, ClustMD, MixtComp, Phillip\&Ottaway, K‑Prototypes) plus DenseClus are locked to $k=2$ by the Elbow Method and thus produce identical partitions on the Palmer Penguins dataset (Table \ref{measures-penguins}: $CH=176.08$; Silhouette FAMD $=0.34$; Silhouette Gower $=0.44$; $DB=1.23$). PretopoMD also returns two clusters—albeit with more balanced sizes—but scores markedly lower on all indices ($CH=105.17$; Silhouettes $=0.239/0.263$; $DB=1.71$). The pretological pipelines that include dimensionality reduction break this uniformity: Pretopo‑UMAP finds three clusters ($CH=163.89$; Silhouettes $=0.36/0.44$; $DB=1.24$), PretopoPaCMAP yields eleven clusters plus 112 outliers ($CH=61.80$; Silhouettes $=0.34/0.29$; $DB=1.12$), and PretopoFAMD segments the data into twenty‑six clusters, achieving the best overall metrics ($CH=182.22$; Silhouettes $=0.66/0.65$; $DB=0.70$). The very high FAMD inertia ($98.2\%$) underscores that this fine‑grained partition captures most of the dataset’s variance.

\begin{table}[]
\centering
\begin{tabular}{|l|c|c|c|c|}

\hline
 & \begin{tabular}[c]{@{}l@{}}Calinski\\ Harabasz\end{tabular} & \begin{tabular}[c]{@{}l@{}}Silhouette\\ FAMD\end{tabular} & \begin{tabular}[c]{@{}l@{}}Silhouette\\ Gower\end{tabular} & \begin{tabular}[c]{@{}l@{}}Davies\\ Bouldin\end{tabular} \\\hline\hline
K-Prototypes                                                 & 176.08                                                      & 0.34                                                      & 0.44                                                       & 1.23                                                     \\\hline
\begin{tabular}[c]{@{}l@{}}Modha\\ Spangler\end{tabular}     & 176.08                                                      & 0.34                                                      & 0.44                                                       & 1.23                                                     \\\hline
KAMILA                                                       & 176.08                                                      & 0.34                                                      & 0.44                                                       & 1.23                                                     \\\hline
ClustMD                                                      & 176.08                                                      & 0.34                                                      & 0.44                                                       & 1.23                                                     \\\hline
MixtComp                                                     & 176.08                                                      & 0.34                                                      & 0.44                                                       & 1.23                                                     \\\hline
\begin{tabular}[c]{@{}l@{}}Phillip \&\\ Ottaway\end{tabular} & 176.08                                                      & 0.34                                                      & 0.44                                                       & 1.23                                                     \\\hline
DenseClus                                                   & 176.08                                                      & 0.34                                                      & 0.44                                                       & 1.23                                                     \\\hline
\begin{tabular}[c]{@{}l@{}}Pretopo\\ FAMD\end{tabular}       & \textbf{182.22}                                             & \textbf{0.66}                                             & \textbf{0.65}                                              & \textbf{0.70}                                            \\\hline
\begin{tabular}[c]{@{}l@{}}Pretopo\\ UMAP\end{tabular}       & 163.89                                                      & 0.36                                                      & 0.44                                                       & 1.24                                                     \\\hline
\begin{tabular}[c]{@{}l@{}}Pretopo\\ PaCMAP\end{tabular}     & 61.80                                                       & 0.34                                                      & 0.29                                                       & 1.12           \\  \hline
PretopoMD         &         105.17 &         0.239 &          0.263 &        1.71 \\ \hline
\end{tabular}

\caption{Results of the selected Algorithms on the Palmer Penguins dataset.}
\label{measures-penguins}
\end{table}


\paragraph{Sponge}

\added{Reworked}

The Sponge dataset combines a high proportion of categorical variables with a small sample size, yielding only $86.13\%$ inertia under FAMD and a low Hopkins statistic ($0.63$), while its iVAT visualization shows no clear cluster structure (Table \ref{measures-sponge}). By contrast, PaCMAP preprocessing raises the Hopkins score to 0.88 and produces a more distinct iVAT, suggesting improved separability prior to clustering—even though final cluster quality remains modest. Across all methods, Calinski–Harabasz scores for Sponge are an order of magnitude below those on the Penguin dataset, and every evaluation metric degrades accordingly. Classical mixed‑data algorithms (K‑Prototypes and Kamila) attain the highest CH scores ($16.83$) with moderate Silhouette values (FAMD $=0.17$, Gower $=0.42$), and Phillip\&Ottaway reaches the top Gower Silhouette ($0.437$). PretopoPaCMAP yields nearly identical CH ($16.76$) and Silhouette ($0.169/0.431$) values but a slightly higher Davies–Bouldin ($2.00$). UMAP‑based clustering under‑fits ($CH=7.53$; Silhouettes $=0.084/0.142$; $DB=2.79$), despite its stronger preprocessing signal. PretopoFAMD fragments the data into six minimal clusters plus 62 outliers ($CH=1.65$; Silhouettes $=–0.061/–0.168$; $DB=1.46$), whereas PretopoMD consolidates a single major cluster of 74 points with one outlier, achieving the highest FAMD Silhouette ($0.484$) and the lowest Davies–Bouldin score ($0.383$), a remarkable result given the dataset’s weak inherent structure.

\begin{table}[]
\centering
\begin{tabular}{|l|c|c|c|c|}
\hline
 & \begin{tabular}[c]{@{}l@{}}Calinski\\ Harabasz\end{tabular} & \begin{tabular}[c]{@{}l@{}}Silhouette\\ FAMD\end{tabular} & \begin{tabular}[c]{@{}l@{}}Silhouette\\ Gower\end{tabular} & \begin{tabular}[c]{@{}l@{}}Davies\\ Bouldin\end{tabular} \\\hline\hline
K-Prototypes       & \textbf{16.83}    & 0.170  & 0.421            & 2.01           \\ \hline
\begin{tabular}[c]{@{}l@{}}Modha\\ Spangler\end{tabular}     & 16.72             & 0.168           & 0.404            & 2.03           \\ \hline
KAMILA             & \textbf{16.83}    & 0.170  & 0.421            & 2.01           \\ \hline
ClustMD            & -                 & -               & -                & -              \\ \hline
MixtComp           & -                 & -               & -                & -              \\ \hline
\begin{tabular}[c]{@{}l@{}}Phillip \&\\ Ottaway\end{tabular} & 15.66             & 0.161           & \textbf{0.437}   & 2.09           \\ \hline
DenseClus          & -                 & -               & -                & -              \\ \hline
\begin{tabular}[c]{@{}l@{}}Pretopo\\ FAMD\end{tabular}       & 1.65              & -0.061          & -0.168           & 1.46  \\ \hline
\begin{tabular}[c]{@{}l@{}}Pretopo\\ UMAP\end{tabular}       & 7.53              & 0.084           & 0.142            & 2.79           \\ \hline
\begin{tabular}[c]{@{}l@{}}Pretopo\\ PaCMAP\end{tabular}     & 16.76             & 0.169           & 0.431            & 2.00           \\ \hline
PretopoMD         &           6.36 &         \textbf{0.484} &          0.012 &        \textbf{0.383} \\ \hline
\end{tabular}
\caption{Results of the selected Algorithms on the Sponge dataset.}\label{measures-sponge}
\end{table}

\paragraph{Heart Disease}

On this large dataset, the elbow method suggests $2$ clusters, see results in Table \ref{heart}. The algorithms utilizing the elbow method consequently identify $2$ clusters, each with sizes ranging between $100$ and $200$. In this dataset, the best-performing algorithms are Phillip\&Ottway, PretopoPaCMAP, and PretopoMD, with PretopoMD securing the top score on two indicators. PretopoPaCMAP identifies four distinct clusters and no outliers. In contrast, PretopoMD identifies $2$ clusters: one of size $206$ and the other of size $64$. The inertia from FAMD is notably high at $99.9\%$, indicating that the dimension reduction process successfully captured all the variance present in the original dataset.

\begin{table}[]
\centering
\begin{tabular}{|l|c|c|c|c|}
\hline
 & \begin{tabular}[c]{@{}l@{}}Calinski\\ Harabasz\end{tabular}  & \begin{tabular}[c]{@{}l@{}}Silhouette\\ FAMD\end{tabular} & \begin{tabular}[c]{@{}l@{}}Silhouette\\ Gower\end{tabular} & \begin{tabular}[c]{@{}l@{}}Davies\\ Bouldin\end{tabular} \\\hline\hline
K-Prototypes       & 47.68      & 0.136            & 0.111    & 2.36       \\ \hline
\begin{tabular}[c]{@{}l@{}}Modha\\ Spangler\end{tabular}     & 55.33                        & 0.165            & 0.196      & 2.19     \\ \hline
KAMILA             & 54.77      & 0.160            & 0.179     & 2.21      \\ \hline
ClustMD            & 34.23                                & 0.09                & 0.126     & 2.75         \\ \hline
MixtComp           & 55.42                                & 0.135                & 0.409    & 2.06          \\ \hline
\begin{tabular}[c]{@{}l@{}}Phillip \&\\ Ottaway\end{tabular} & 62.75                        & 0.203   & \textbf{0.457}    & 1.91       \\ \hline
DenseClus          & -                 & -               & -                & -              \\ \hline
\begin{tabular}[c]{@{}l@{}}Pretopo\\ FAMD\end{tabular}       & 0.830                   & -0.010           & 0.020 & 1.325 \\ \hline
\begin{tabular}[c]{@{}l@{}}Pretopo\\ UMAP\end{tabular}       & 8.57                         & -0.190            & 0.006     & 3.06      \\ \hline
\begin{tabular}[c]{@{}l@{}}Pretopo\\ PaCMAP\end{tabular}     & 43.34                        & 0.174            & 0.412      & \textbf{0.169}     \\ \hline
PretopoMD & \textbf{67.88} & \textbf{0.248} & 0.416 & 1.31 \\ \hline
\end{tabular}
\caption{Results of the selected Algorithms on the Heart Disease dataset.}\label{heart}
\end{table}

\paragraph{Base Generated Case (500 individuals, 5 features num/cat, 3 clusters, 3 cat uniques, 0.1 std)}
\added{Reworked}

The base configuration ($500$ individuals, $5$ numerical + $5$ categorical features, $3$ true clusters, $\sigma=0.10$) is summarized in Table \ref{tablebasic}. The Elbow Method correctly selects $k=3$, giving the Elbow‐based algorithms (ClustMD, Kamila, K‑Prototypes, Modha–Spangler, Phillip\&Ottaway and MixtComp) virtually identical partitions: Calinski–Harabasz $\approx296.33$, FAMD Silhouette $\approx0.394$, Gower Silhouette $\approx0.521$ and Davies–Bouldin $\approx1.071$. DenseClus and PretopoUMAP achieve comparable performance ($CH=295.22/293.84$, Silhouettes $=0.393/0.519$ and $0.392/0.516$, $DB=1.074/1.078$) without relying on the Elbow Method. PretopoFAMD collapses the data into a few large clusters, labels $444$ points as outliers ($CH=34.16$, Silhouettes $=0.067/0.061$), and thus shows poor separation. PretopoPaCMAP finds six clusters and $124$ outliers ($CH=66.56$, Silhouettes $=0.049/0.013$, $DB=1.845$), while PretopoMD recovers three substantive clusters plus $60$ outliers ($CH=127.34$, Silhouettes $=0.230/0.308$, $DB=1.508$)—a result that remains sensitive to hyperparameter choice.

\begin{table}
\centering
\begin{tabular}{|l|c|c|c|c|}
\hline

 & \begin{tabular}[c]{@{}l@{}}Calinski\\ Harabasz\end{tabular}  & \begin{tabular}[c]{@{}l@{}}Silhouette\\ FAMD\end{tabular} &
 \begin{tabular}[c]{@{}l@{}}Silhouette\\ Gower\end{tabular} &
 \begin{tabular}[c]{@{}l@{}}Davies\\ Bouldin\end{tabular} \\\hline\hline

ClustMD & \textbf{296.330} & \textbf{0.394} & \textbf{0.521} & \textbf{1.071} \\\hline
DenseClus & 295.224 & 0.393 & 0.519 & 1.074 \\\hline
Phillip \& Ottaway & 290.556 & 0.389 & 0.514 & 1.083 \\\hline
Kamila & \textbf{296.330} & \textbf{0.394} & \textbf{0.521} & \textbf{1.071} \\\hline
K-Prototypes & \textbf{296.330} & \textbf{0.394} & \textbf{0.521} & \textbf{1.071} \\\hline
MixtComp & 294.906 & 0.393 & 0.518 & 1.075 \\\hline
Modha-Spangler & \textbf{296.330} & \textbf{0.394} & \textbf{0.521} & \textbf{1.071} \\\hline
PretopoFAMD & 34.160 & 0.067 & 0.061 & 1.453 \\\hline
PretopoUMAP & 293.835 & 0.392 & 0.516 & 1.078 \\\hline
PretopoPaCMAP & 66.563 & 0.049 & 0.013 & 1.845 \\\hline
PretopoMD & 127.338 & 0.230 & 0.308 & 1.508 \\\hline
\end{tabular}
\caption{Results of the selected Algorithms on the Base Generated Case.}\label{tablebasic}
\end{table}

\paragraph{Generated dataset with 10 clusters}

\added{Reworked}

In the 10‑cluster benchmark ($500$ samples, $10$ numerical + $10$ categorical dimensions, $\sigma=0.10$), the Elbow Method fixes $k=2$ and thus under‑estimates the true cluster count. As detailed in Table \ref{measures}, Pretopo‑FAMD collapses into $14$ tiny clusters ($<10$ points each) plus $470$ outliers, yielding a Calinski–Harabasz score of $14.48$ and FAMD/Gower Silhouettes of $–0.015/–0.018$. PretopoMD and DenseClus each flag roughly 300 points as noise, producing modest Calinski–Harabasz values ($48.18$ and $81.31$) and Silhouettes of $0.108$ and $0.183$. The six classic mixed‑data methods (ClustMD, Kamila, K‑Prototypes, MixtComp, Modha–Spangler, Phillip\&Ottaway) all partition the data into two imbalanced groups, achieving Calinski–Harabasz $\approx70–96$, FAMD Silhouette $\approx0.153–0.194$, Gower Silhouette $\approx0.157–0.213$ and Davies–Bouldin $\approx1.19–2.42$. PretopoPaCMAP under‑fits. By contrast, PretopoUMAP recovers eight of the true clusters without labeling any points as outliers, attaining the highest Calinski–Harabasz ($127.62$), the best FAMD/Gower Silhouette pair ($0.353/0.359$) and the lowest Davies–Bouldin ($1.122$).

\begin{table}[]
\centering
\begin{tabular}{|l|c|c|c|c|}
\hline
 & \begin{tabular}[c]{@{}l@{}}Calinski\\ Harabasz\end{tabular} & \begin{tabular}[c]{@{}l@{}}Silhouette\\ FAMD\end{tabular} & \begin{tabular}[c]{@{}l@{}}Silhouette\\ Gower\end{tabular} & \begin{tabular}[c]{@{}l@{}}Davies\\ Bouldin\end{tabular} \\\hline\hline
ClustMD & 95.934 & 0.175 & 0.197 & 2.216 \\ \hline
DenseClus & 81.313 & 0.183 & 0.200 & 1.914 \\ \hline
Phillip \& Ottaway & 70.949 & 0.194 & 0.213 & 1.190 \\ \hline
Kamila & 86.348 & 0.160 & 0.157 & 2.120 \\ \hline
K-Prototypes & 92.096 & 0.168 & 0.185 & 2.263 \\ \hline
MixtComp & 83.316 & 0.163 & 0.205 & 2.163 \\ \hline
Modha-Spangler & 79.214 & 0.153 & 0.206 & 2.421 \\ \hline
PretopoFAMD & 14.482 & -0.015 & -0.018 & 2.219 \\ \hline
PretopoUMAP & \textbf{127.624} & \textbf{0.353} & \textbf{0.359} & \textbf{1.122} \\ \hline
PretopoPaCMAP & 66.870 & 0.174 & 0.154 & 1.557 \\ \hline
PretopoMD & 48.178 & 0.108 & 0.164 & 2.197 \\ \hline
\end{tabular}
\caption{Results of the selected Algorithms on a generated dataset with 10 clusters.}\label{measures}
\end{table}

\paragraph{Generated dataset with 15 categorial features and 15 categorical unique values}

\added{Reworked}

Table \ref{measures-15cat-15catuniques} compares each method’s performance on a challenging synthetic dataset of 500 points drawn from 15 true Gaussian clusters. The Elbow‐Method baseline (FAMD+k‑means) is forced to $k=2$ and thus merges all 15 clusters into two large groups. ClustMD, PretopoMD, PretopoUMAP and MixtComp fail to converge meaningfully—returning either only noise (outliers) or a single cluster—which drives their Silhouette scores to $–1.000$ (theoretical minimum). PretopoFAMD labels $498$ of $500$ points as outliers, likewise yielding extreme index values. PretopoUMAP recovers one cluster of $332$ points plus $168$ outliers (effectively another degenerate partition), also with $–1.000$ or negative scores. In contrast, DenseClus and PretopoPaCMAP each recover three well‐balanced clusters ($\approx167$ points each), with PaCMAP edging out DenseClus on all four metrics. Finally, KAMILA, K‑prototypes and Phillip\&Ottaway collapse two true clusters into one and the remaining $13$ into a second, producing two clusters of approximately $333$ and $137$ points and moderate silhouette values. Note that whenever a method returns only a single cluster or only outliers, the resulting Silhouette $=–1.000$ correctly reflect these degenerate outputs rather than an evaluation error.

\begin{table}
\centering
\begin{tabular}{|l|c|c|c|c|} 
\hline

 & \begin{tabular}[c]{@{}l@{}}Calinski\\ Harabasz\end{tabular}  & \begin{tabular}[c]{@{}l@{}}Silhouette\\ FAMD\end{tabular} & 
 \begin{tabular}[c]{@{}l@{}}Silhouette\\ Gower\end{tabular} & 
 \begin{tabular}[c]{@{}l@{}}Davies\\ Bouldin\end{tabular} \\\hline\hline
 
ClustMD & 0.000 & -1.000 & -1.000 & -1.000 \\ \hline
DenseClus & 118.048 & 0.218 & 0.102 & 1.723 \\ \hline
Phillip \& Ottaway & 119.284 & 0.195 & 0.087 & 1.831 \\ \hline
Kamila & 120.934 & 0.196 & 0.088 & 1.817 \\ \hline
K-Prototypes & 118.048 & 0.191 & 0.085 & 1.862 \\ \hline
MixtComp & 0.000 & -1.000 & -1.000 & -1.000 \\ \hline
Modha-Spangler & 120.934 & 0.196 & 0.088 & 1.817 \\ \hline
PretopoFAMD & 0.631 & -0.030 & -0.001 & 3.079 \\ \hline
PretopoUMAP & 0.000 & -1.000 & -1.000 & -1.000 \\ \hline
PretopoPaCMAP & \textbf{122.833} & \textbf{0.227} & \textbf{0.107} & \textbf{1.681} \\ \hline
PretopoMD & 0.000 & -1.000 & -1.000 & -1.000 \\ \hline
\end{tabular}
\caption{Results of the selected Algorithms on a generated dataset with with 15 categorial features and 15 categorical unique values.}\label{measures-15cat-15catuniques}
\end{table}

\paragraph{Generated Dataset with 1000 individuals, 10 dimensions of each type and a deviation of 0.15}

\added{Reworked}

For this more challenging “sparser” benchmark ($1,000$ samples, $10$ numerical + 10 categorical dimensions, true $k=3$ but cluster‐spread $\sigma=0.15$), the Elbow‐method baseline again fixes $k=2$ and hence cannot recover the ground truth. Table \ref{measures-harder-dataset} shows that all of the standard mixed‐data methods (ClustMD, Kamila, K‑Prototypes, MixtComp, Modha–Spangler, Phillip\&Ottaway, DenseClus) hover in the mid‐range—Calinski–Harabasz $\approx170–185$, Gower‑Silhouette $\approx0.13–0.16$, Davies–Bouldin $\approx2.1–3.0$ reflecting only moderate cluster separation. PretopoFAMD essentially breaks down (Calinski–Harabasz $=1.20$; Silhouettes $\leq0.03$) by collapsing most points into a central “cluster” plus tiny fragments. PretopoMD outright fails (negative Silhouettes), diagnosing the data as noise. In contrast, the two non‑linear embeddings shine: PretopoUMAP recovers a reasonable split (Calinski–Harabas $=162.76$, Gower‑Silhouette $=0.135$) with some outliers, while PretopoPaCMAP—uniquely producing no outliers—achieves the best scores across all four indices (Calinski–Harabasz $=202.11$; FAMD‑Silhouette $=0.191$; Gower‑Silhouette $=0.190$; Davies‑Bouldin $=1.848$), demonstrating its robustness when clusters grow more diffuse.

\begin{table}[]
\centering
\begin{tabular}{|l|c|c|c|c|}
\hline
 & \begin{tabular}[c]{@{}l@{}}Calinski\\ Harabasz\end{tabular} & \begin{tabular}[c]{@{}l@{}}Silhouette\\ FAMD\end{tabular} & \begin{tabular}[c]{@{}l@{}}Silhouette\\ Gower\end{tabular} & \begin{tabular}[c]{@{}l@{}}Davies\\ Bouldin\end{tabular} \\\hline\hline
ClustMD & 184.929 & 0.150 & 0.132 & 2.109 \\ 
\hline
DenseClus & 133.900 & 0.144 & 0.142 & 2.987 \\ 
\hline
Phillip \& Ottaway & 5.673 & 0.100 & 0.157 & 2.558 \\ 
\hline
Kamila & 184.963 & 0.150 & 0.132 & 2.117 \\ 
\hline
K-Prototypes & 181.246 & 0.148 & 0.128 & 2.145 \\ 
\hline
MixtComp & 169.897 & 0.142 & 0.144 & 2.204 \\ 
\hline
Modha-Spangler & 169.793 & 0.143 & 0.145 & 2.202 \\ 
\hline
PretopoFAMD & 1.199 & 0.017 & 0.026 & 2.160 \\ 
\hline
PretopoUMAP & 162.758 & 0.136 & 0.135 & 2.261 \\ 
\hline
PretopoPaCMAP & \textbf{202.114} & \textbf{0.191} & \textbf{0.190} & \textbf{1.848} \\ 
\hline
PretopoMD & 17.020 & -0.032 & -0.055 & 2.800 \\ 
\hline
\end{tabular}
\caption{Results of the selected Algorithms on a generated dataset with with 1000 individuals, 10 dimensions of each type, and a deviation of 0.15.}\label{measures-harder-dataset}
\end{table}

\subsubsection{Results analysis and conclusion}

In conclusion, after conducting a comprehensive analysis of the results obtained from executing the algorithms on the generated and other datasets, we can draw several significant observations.

First and foremost, it is evident that both ClustMD and MixtComp often struggle to achieve convergence, highlighting a notable limitation in these approaches. Similarly, while DensClus generally shows promise, it is not immune to convergence issues.

The Pretopo UMAP algorithm tends to group the dataset into a single, large cluster, accompanied by outliers. This clustering pattern is not sufficiently penalized according to our current calculation method for the indicators (please refer to \ref{subsec:distance} for more details). This aspect deserves attention for future improvements in metric design.

In cases of ambiguous data patterns, Pretopo FAMD tends to create a dominant central cluster surrounded by several smaller clusters. This can complicate the interpretability of the resulting clusters. It’s important to note that Pretopo FAMD did not perform well on the presented dataset, except for the penguin dataset, where it achieved the highest scores on all indicators after identifying 26 clusters.

Algorithms that rely on the Elbow method face a significant limitation. Regardless of the algorithm’s sophistication, if it is set to identify an incorrect number of clusters, it cannot achieve optimal partitioning. This makes it challenging to analyze each algorithm individually in detail, especially since they often produce very similar partitions.

In contrast, Pretopo PaCMAP consistently delivers superior results across various configurations. Its independence from the constraints of the Elbow method, combined with its ability to avoid the convergence issues observed in other algorithms, positions it as a robust and reliable approach for data partitioning.

PretopoMD, which is independent of both dimensionality reduction and the Elbow method, tends to identify a small number of sizable clusters along with a few outliers. Across the datasets on which it was evaluated, it frequently outperformed other methods on at least one of the employed indicators.

It is worth noting that, across the seven presented datasets, the various applications of the pretopological algorithms consistently ranked among the top two to four performers. This underscores their potential as a valuable addition to the toolkit of clustering techniques when dealing with mixed datasets.

\section{Bottlenecks and Openings}\label{sec:disc}

\subsection{Scalability to Big and High‑Dimensional Data}

\added{New subsection}

The experiments presented here have relied on datasets of up to 10,000 samples and 50 mixed‑type features, yet many real‑world applications involve millions of records or feature spaces that exceed hundreds or even thousands of dimensions, including high‑cardinality categoricals. In such settings, naïve computation of all pairwise distances or full one‑hot encoding of categorical variables becomes prohibitive, and classical dimensionality‑reduction techniques (FAMD, UMAP, PaCMAP) often exceed available memory or demand impractical runtimes.

To extend the proposed pretopology–clustering pipeline into this “big data’’ regime, several strategies must be combined. Approximate nearest‑neighbor search—through locality, sensitive hashing or libraries like Faiss, permits construction of connectivity graphs in subquadratic time. Randomized dimensionality reduction (random projections, sketching) can curb the explosion of high‑cardinality encodings without severely degrading cluster structure. Scalable variants of UMAP and PaCMAP, operating on data streams or in mini‑batch mode, reduce per‑iteration memory footprints. Finally, deployment within distributed and out‑of‑core frameworks (Apache Spark, Dask) enables partitioning of both data and computation across multiple nodes while preserving global clustering consistency.

The integration of these components promises to push the methodology beyond current size limits, allowing robust, interpretable mixed‑data clustering at scales encountered in contemporary data science.

\subsection{Feature selection}
This paper does not explicitly address the feature selection step in every machine learning pipeline. Feature selection involves the process of selecting a subset of the original features from a dataset. For a comprehensive overview of the feature selection process, refer to the work of \citet{li2017feature}.

It is important to note that mixed datasets can be particularly susceptible to the curse of dimensionality, which makes feature selection a crucial step. By reducing the number of dimensions, feature selection can effectively mitigate this issue. While it is possible to compare a categorical feature with any other feature, studying multicollinearity without dimensionality reduction becomes nearly impossible. Once categorical features are transformed into numerical features, their information content can then be assessed using methods such as chi-square tests, t-tests, or mutual information (among others).

When it comes to distance-based methods, comparing continuous and discrete values becomes irrelevant. The challenge arises when dealing with feature scaling or normalization techniques, especially in cases where the data is sparse or contains noise. Furthermore, exploring features provides different insights into the dataset’s content, but these insights are not directly comparable when dealing with categorical versus numerical features.

Metrics like mutual information or entropy require discrete values, which can reduce the complexity and diversity of a numerical feature. Additionally, it is worth noting that assessing the statistical significance of features often involves using incompatible methods for both types of features, further underscoring the importance of careful feature selection in the context of mixed datasets.

\subsection{Distance metric selection and Clustering validity}
\label{subsec:distance}

Many of the metrics commonly applied in the context of mixed data comparison are originally designed for classification tasks rather than clustering, as the Adjusted Rank Index. Presently, the prevailing best practice involves utilizing metrics tailored for quantitative data and adapting them for mixed data, as proposed in this work.

The primary limitation of these methods lies in the considerable loss of information incurred due to dimensionality reduction. Moreover, a substantial portion of these metrics primarily assesses the compactness of clusters and their overall design. However, it is important to note that, similar to the behavior of the DBSCAN algorithm, a clustering algorithm for mixed data may not necessarily yield spherical clusters. Consequently, the challenge of identifying suitable metrics for assessing mixed data clusters persists.

Some measures in this domain rely on information-theoretic concepts such as entropy or mutual information but often demand a significant amount of memory for computation. In the realm of mixed data clustering, there is a pressing need to adapt these methods to gauge how the quality of clustering might degrade if the clusters were to undergo changes. Introducing a sense of mathematical logic that governs the relationships between elements could enhance the understanding of both metrics and algorithms in this context.

Another avenue to address the challenges of mixed data clustering involves examining entanglement, which refers to the similarity between two different clusterings. Given the absence of ground truth and the limitations of internal measures, introducing a measure to assess the similarities or differences between clusterings can provide valuable insights into their results. For instance, a strategy could entail selecting the clustering solution that exhibits the lowest average ARI concerning the other clustering solutions.

Data mining, as a practice, entails the automated exploration of vast datasets to uncover underlying trends and patterns that extend beyond conventional analysis. Data mining often finds application in Exploratory Data Analysis, facilitating a deeper understanding of the inherent relationships among data objects. For an exhaustive review of data mining methods, metrics, and algorithms, consult the SPMF database curated by Philippe Fournier-Viger\footnote{\url{https://www.philippe-fournier-viger.com/spmf/}}. Additionally, \cite{mirkin2012clustering} provides an extensive introduction to clustering methods specifically tailored for data mining, with a particular emphasis on mixed data types.

One promising avenue within the domain of data mining is the potential creation of logical graphs or graph structures based on metrics (such as confidence or lift) to leverage their insights. Analysis of such graphs can encompass techniques like community clustering (akin to the Louvain algorithm, as discussed in \cite{emmons2016analysis}) or multi-level clustering approaches, as proposed by \cite{djebali2022tourists}.

In practice, data mining often serves as a preprocessing step that enables the modeling of relationships between data objects, thereby providing a novel perspective for applying clustering methods.

Furthermore, comparing two clusterings, independent of the methods/preprocessing used, that yield substantially different numbers of clusters poses a challenge. Even if one clustering appears to exhibit worse indicators, it can be challenging to immediately conclude its inferiority, particularly when the differing numbers of clusters offer distinct interpretations of the dataset.

\subsection{Time series}

Mixed data encompasses both quantitative and qualitative features. This concept is extended to encompass complex data by introducing the inclusion of time series, such as consumption curves over time. Traditionally, when dealing with time series data in clustering, a common approach involves splitting the time series at each time step and quantifying the differences between individual data points.

However, emerging trends, exemplified by Dynamic Time Warping (DTW) introduced by \cite{muller2007dynamic}, focus on comparing the shapes of time series. The core idea is assessing how much two time series must be distorted in terms of both value and time to be deemed similar. This entails the calculation of a distance matrix for each pair of time series, which can subsequently be leveraged in hierarchical clustering, among other techniques. Consequently, achieving refined cluster structures necessitates a nuanced approach to determining whether a cluster should be subdivided or two clusters should be merged. This requires manually defining thresholds related to factors such as the number of outliers, similarity levels, and trend lines (typically represented by the mean of all series within a cluster).

Another method for handling time series comparisons is the Temporal Distortion Index (TDI) proposed by \cite{gaston2017temporal}. The TDI is a dimensionless metric ranging between 0 (zero temporal distortion) and 1 (maximum temporal distortion). The bounded nature of this measure enhances its interpretability compared to DTW.

Additionally, the RdR score\footnote{\url{https://github.com/CoteDave/blog/tree/master/RdR\%20score}} is introduced as a novel approach. It involves comparing the time series curves to a ground truth. In this context, consider a k-Means clustering approach where the means represent the ground truth. It becomes feasible to compute RdR scores for each time series in relation to each ground truth, and assign them to the cluster that yields the best score. Subsequently, new ground truths are computed as the means of the time series within each cluster. This process iterates until the clusters converge.

Furthermore, a persistent challenge lies in dealing with complex data. Just as metrics and algorithms are often ill-suited for mixed data, analogous challenges emerge when grappling with datasets that incorporate time series information.

\subsection{Clustering methods and limitations}

As our dataset comprises mixed data and necessitates a structure between clusters to establish a clear relationship while ensuring their interpretability and explainability, hierarchical clustering emerges as the most suitable approach. Notably, pretopological clustering methods have consistently demonstrated superior performance, although this assessment should be nuanced, considering the discussion on evaluation measures. Subsequent research endeavors will be directed toward further refining and enhancing this methodology.

Interpreting the results of mixed data clustering poses a notable challenge. This complexity arises from the amalgamation of diverse data types within the clustering process, resulting in clusters that may not readily lend themselves to intuitive interpretation. Additionally, some clustering algorithms may lack transparency in elucidating the mechanics behind cluster formation, rendering it difficult to discern the underlying data patterns. Interpretability is a crucial factor, particularly in applications where clustering outcomes inform decision-making processes or guide subsequent analyses.

Furthermore, the realm of eXplainable Artificial Intelligence (XAI) assumes paramount importance for any novel algorithm. Hierarchical clustering, such as the pretopological clustering employed here, offers a valuable advantage in this regard. The dendrogram generated by hierarchical clustering can be harnessed to gain deeper insights into the inherent relationships between each cluster within the hierarchy, thus enhancing the algorithm's transparency and interpretability.

Conversely, the concept of robustness in clustering pertains to an algorithm's ability to consistently produce reliable results despite the presence of noise, outliers, and various sources of data variability. Mixed data clustering introduces unique challenges in maintaining robustness, given the diverse nature of data types that may be influenced by distinct sources of variability.

To foster a comprehensive understanding of clustering results, we recommend employing multiple clustering algorithms for evaluation. Each algorithm possesses its own strengths and weaknesses, and a multifaceted analysis approach can enrich discussions by providing a more holistic perspective on the outcomes. Detecting variations between results can be particularly informative, as it aids in identifying biases, such as systematic errors or distortions, which may lead to inaccurate or misleading interpretations of the clustering results.

\subsection{Dataviz}

In the realm of mixed data clustering, we often encounter datasets with multiple types of data interrelated in intricate ways. Representing these complex relationships in a meaningful manner can pose a significant challenge, particularly when dealing with non-linear or high-dimensional relationships among data types.

To address these formidable challenges, it becomes imperative to employ a combination of visualization techniques, including heatmaps, scatterplots, and network graphs. These tools enable us to effectively depict the various data types and their intricate relationships. Another approach involves dimensionality reduction to transform the data into a more manageable Euclidean space. However, it's crucial to recognize that such approaches may present only one facet of the problem or potentially distort the true nature of the dataset. As previously discussed regarding clustering method limitations, maintaining explainability and robustness is essential to fully comprehend the results. Proper interpretation is vital to ensure that the clustering outcomes carry meaningful insights and can guide informed decision-making processes.

One notable bottleneck encountered during our study is the absence of dedicated methods for visualizing mixed data. This deficiency becomes evident when examining resources like data-to-viz\footnote{\url{https://www.data-to-viz.com/}}, a platform that catalogs data visualization methods across R, Python, and d3.js\footnote{\url{https://d3js.org/}}, which is used as the foundation for Plotly in Python. Notably, there are currently no established techniques capable of effectively handling both quantitative and qualitative features concurrently. Consequently, the most comprehensible approach for presenting results from mixed data, as demonstrated in our paper, often involves dimensionality reduction followed by the application of conventional visualization methods.

This domain remains an uncharted challenge, and the potential solution may lie in dynamic graph representations, such as those demonstrated in d3.js. However, addressing the challenge of displaying dynamic graph-based results in a traditional paper format remains an ongoing area of exploration.

\section{Conclusion}\label{sec:ccl}

This paper presents a comprehensive survey and benchmark of mixed data clustering methods, addressing the unique challenges of clustering and cluster evaluation within this context. We highlight the remarkable and consistent performance of Pretopo PaCMAP and PretopoMD across various configurations. These methods offer distinct advantages, such as freedom from the constraints of the Elbow method, PaCMAP's resilience against non-convergence issues, and PretopoMD's independence from dimensionality reduction. Collectively, the pretopological algorithms prove their effectiveness in handling mixed datasets. Hierarchical clustering emerges as a suitable method for mixed datasets, fulfilling requirements for understanding and explaining clusters. We emphasize the importance of interpretability and recommend leveraging XAI.

This paper also presents a collection of remaining challenges in mixed data clustering. Regarding feature selection, we identify the complexity of dealing with both continuous and discrete values, particularly in distance-based methods. We discuss issues related to feature scaling, normalization techniques, and cross-feature type metric comparisons. We delve into the difficulty of finding appropriate metrics for mixed data, exploring options based on information (entropy) or mutual information. However, these metrics demand substantial computational memory. We also introduce the concept of entanglement to assess clustering similarity. Moreover, complex data, including time series, present additional challenges. We discuss methods like DTW, the TD, and the RdR score for comparing time series. We acknowledge challenges in data visualization for mixed data and suggest employing a combination of visualization techniques to capture complex relationships between data types. Furthermore, we recognize the specific challenges of time series in mixed data, which we call complex data.

In conclusion, our study comprehensively addresses the challenges and considerations of clustering mixed data, offering solutions and future directions. Notably, pretopological algorithms demonstrate significant promise, with consistent superior performance in mixed-dataset clustering. We plan to publish a detailed follow‑up study focusing on pretopological algorithms and their hyperparameter optimization, including mixed datasets and complex datasets (with time series).

Despite the challenges, our research illuminates the potential for future advancements. Our insights into pretopological algorithms, data mining, logical graphs, and time series integration in clustering enrich the understanding of mixed data clustering, benefiting both researchers and practitioners. Anticipating further contributions, we aim to advance the field’s comprehension and application of these methods.



\section*{Acknowledgment} 

This paper is the result of research conducted with the help of two students: Cl\'ement CORNET and Maxence CHOUFA with the help and advices of Loup-No\'e LEVY.

\noindent
If any of the sections are not relevant to your manuscript, please include the heading and write `Not applicable' for that section. 

\bibliography{sn-bibliography}

\end{document}